%% file: main.tex
\RequirePackage{fix-cm}
\documentclass[twocolumn]{svjour3}          
\smartqed  
\usepackage{graphicx}
\usepackage{import}
\usepackage[numbers]{natbib}
\usepackage{times}
\usepackage{epsfig}
\usepackage{amsmath}
\usepackage{amssymb}
\usepackage{soul}
\usepackage{dsfont}
\usepackage{multirow}
\usepackage{capt-of}
\usepackage{graphics}
\usepackage{algorithm}
\usepackage{algorithmic}
\usepackage{dsfont}
\usepackage{booktabs} 

\usepackage{color, colortbl}
\definecolor{Gray}{gray}{0.9}
\newcolumntype{g}{>{\columncolor{Gray}}c}
\usepackage[dvipsnames]{xcolor}
\usepackage{hyperref}

\usepackage{flexisym}
\usepackage{ifthen}
\usepackage{color}
\usepackage{bigstrut}
\usepackage{wasysym}

\input{macros.tex}

\begin{document}

\title{Manhattan Room Layout Reconstruction from a Single 360$^{\circ}$ image: A Comparative Study of State-of-the-art Methods}





\author{Chuhang Zou$^{\ast1}$ \and
        Jheng-Wei Su$^{\ast2}$ \and
        Chi-Han Peng$^{3,4}$ \and
        Alex Colburn$^5$ \and
        Qi Shan$^{\star6}$ \and
        Peter Wonka$^7$ \and
        Hung-Kuo Chu$^2$ \and
        Derek Hoiem$^1$
}


\institute{
$^\ast$ Equal contribution\\
$^\star$ This work is not done while at apple\\
Chuhang Zou, czou4@illinois.edu\\
Jheng-Wei Su, jhengweisu@cloud.nthu.edu.tw\\
Chi-Han Peng, pchihan@asu.edu\\
Alex Colburn, alex@colburn.org\\
Qi Shan, qshan@apple.com\\
Peter Wonka, pwonka@gmail.com\\
Hung-Kuo Chu, hkchu@cs.nthu.edu.tw\\
Derek Hoiem, dhoiem@illinois.edu\\
$^1$ University of Illinois at Urbana-Champaign\\
$^2$ National Tsing Hua University\\
$^3$ National Chiao Tung University\\
$^4$ ShanghaiTech University\\
$^5$ University of Washington\\
$^6$ Apple Inc.\\
$^7$ King Abdullah University of Science and Technology
}

\date{Received: date / Accepted: date}

\maketitle

\begin{abstract}
Recent approaches for predicting layouts from 360$^{\circ}$ panoramas produce excellent results. These approaches build on a common framework consisting of three steps: a pre-processing step based on edge-based alignment, prediction of layout elements, and a post-processing step by fitting a 3D layout to the layout elements.  Until now, it has been difficult to compare the methods due to multiple different design decisions, such as the encoding network~(\eg SegNet or ResNet), type of elements predicted (\eg corners, wall/floor boundaries, or semantic segmentation), or method of fitting the 3D layout.  To address this challenge, we summarize and describe the common framework, the variants, and the impact of the design decisions. For a complete evaluation, we also propose extended annotations for the Matterport3D dataset~\cite{chang2017matterport3d}, and introduce two depth-based evaluation metrics.

\end{abstract}

\input{Introduction.tex}

\input{Related_work.tex}

\input{Dataset.tex}

\input{Matterport3D.tex}

\input{Method_overview.tex}

\input{Performance_comparison.tex}

\input{Conclusion.tex}

\begin{acknowledgements}
This research is supported in part by ONR MURI grant N00014-16-1-2007, iStaging Corp. fund and the Ministry of Science and Technology of Taiwan (108-2218-E-007-050- and 107-2221-E-007-088-MY3).
We thank Shang-Ta Yang for providing the source code of DuLa-Net. We thank Cheng Sun for providing the source code of HorizonNet and help run experiments on our provided dataset.
\end{acknowledgements}

\bibliographystyle{spbasic}      
\bibliography{egbib}   

\end{document}

%% file: macros.tex
\def\etal{et~al.}			  
\def\eg{e.g.,~}               
\def\ie{i.e.,~}               
\def\vs{vs.~}                 

\newcommand{\secref}[1]{Sec.~\ref{#1}}
\newcommand{\figref}[1]{Fig.~\ref{#1}} 
\newcommand{\tabref}[1]{Table~\ref{#1}}
\newcommand{\eqnref}[1]{Eqn.~\ref{#1}}

\newcommand{\layoutnetEX}{LayoutNet v2}
\newcommand{\dulanetEX}{DuLa-Net v2}
\newcommand{\datasetName}{MatterportLayout}

\newcommand{\EtoP}{E2P}

\newcommand{\fcmap}{M_{FC}}
\newcommand{\fpmap}{M_{FP}}
\newcommand{\fcName}{floor-ceiling probability map}
\newcommand{\fpName}{floor plan probability map}
\newcommand{\ffpName}{fused floor plan probability map}
\newcommand{\fcmapCeil}{\fcmap^{C}}
\newcommand{\fcmapFloor}{\fcmap^{F}}

\newcommand{\ceilBranch}{ceiling-branch}
\newcommand{\panoBranch}{panorama-branch}

\newcommand\etc{etc\@ifnextchar.{}{.\@}}

%% file: Introduction.tex
\input{fig_illustration.tex}

\section{Introduction}
\label{sec:intro}

Estimating the 3D room layout of indoor environment is an important step toward a holistic scene understanding and would benefit many applications such as robotics and virtual/augmented reality. The room layout specifies the positions, orientations, and heights of the walls, relative to the camera center. The layout can be represented as a set of projected corner positions or boundaries or as a 3D mesh. Existing works apply to special cases of the problem such as predicting cuboid-shaped layouts from perspective images or from panoramic images.

Recently, various approaches~\cite{zou2018layoutnet,yang2019dula,sun2019horizonnet} for 3D room layout reconstruction from a single panoramic image have been proposed, which all produce excellent results. These methods are not only able to reconstruct cuboid room shapes, but also estimate non-cuboid general Manhattan layouts as shown in~\figref{fig:illustration}. Different from previous work~\cite{zhang2014panocontext} that estimates 3D layouts by decomposing a panorama into perspective images, these approaches operate directly on the panoramic image in equirectangular view, which effectively reduces the inference time. These methods all follow a common framework: (1) a pre-processing edge-based alignment step, ensuring that wall-wall boundaries are vertical lines and substantially reducing prediction error; (2) a deep neural network that predicts the layout elements, such as layout boundaries and corner positions~(LayoutNet~\cite{zou2018layoutnet} and HorizonNet~\cite{sun2019horizonnet}), or a semantic 2D floor plan in the ceiling view~(DuLa-Net~\cite{yang2019dula}); and (3) a post-processing step that fits the~(Manhattan) 3D layout to the predicted elements.

However, until now, it has been difficult to compare these methods due to multiple different design decisions. For example, LayoutNet uses SegNet as encoder while DuLa-Net and HorizonNet use ResNet; HorizonNet applies random stretching data augmentation~\cite{sun2019horizonnet} in training, while LayoutNet and DuLa-Net do not. Direct comparison of the three methods may conflate impact of contributions and design decisions. We therefore want to isolate the effects of the contributions by comparing performance with the same encoding architectures and other settings. 
Moreover, 
given the same network prediction, we want to compare the performance by using different post-processing steps~(under equirectangular view or ceiling view). 
Therefore, in this paper, the authors of LayoutNet and DuLa-Net work together to better describe the common framework, the variants, and the impact of the design decisions for 3D layout estimation from a single panoramic image. For a detailed comparison, we evaluate performance using a unified encoder (\ie ResNet~\cite{he2016deep}) and consistent training details such as random stretching data augmentation, and discuss effects using different post-processing steps. 
Based on the modifications to LayoutNet and DuLa-Net listed above, we propose the improved version called {\layoutnetEX}\footnote{Code is available at: \url{https://github.com/zouchuhang/LayoutNetv2}} and {\dulanetEX}\footnote{Code is available at: \url{https://github.com/SunDaDenny/DuLa-Net}}, which achieve the state-of-the-art for cuboid layout reconstruction.

To compare performance for reconstructing different types of 3D room shape, we extend the annotations of Matterport3D dataset with ground truth 3D layouts\footnote{Our annotation is available at: \url{ https://github.com/ericsujw/Matterport3DLayoutAnnotation}}. Unlike existing public datasets, such as the PanoContext dataset~\cite{zhang2014panocontext}, which provides mostly cuboid layouts of only two scene types,~\ie bedroom and living room, the Matterport3D dataset contains 2,295 real indoor RGB-D panoramas of more than 12 scene types,~\eg kitchen, office, and corridor, and 10 general 3D layouts,~\eg ``L''-shape and ``T''-shape. 
Moreover, we leverage the depth channel of dataset images and introduce two depth-based evaluation metrics for comparing general Manhattan layout reconstruction performance. Depth-based metrics are more general than 3D IoU or 2D pixel labeling error, as they pertain to the geometric error of arbitrary-shaped scenes. Additionally, depth-based metrics could be used to compare layout algorithms in the context of multi-view depth and scene reconstruction. We show in experiments that when the performance gap between the competing methods is small in 2D and 3D IoU metrics, the performance gap in the depth-based metrics is still more distinguishable, which helps to provide fruitful comparisons for quantitative analysis. 

The experimental results demonstrate that: (1) LayoutNet's decoder can better capture global room shape context, performing the best for cuboid layout reconstruction and being robust to foreground occlusions. 
(2) For non-cuboid layout estimation, DuLa-Net and HorizonNet's decoder can better capture detailed layout shapes like pipes, showing better generalization to various complex layout types. Their simplified network output representation also take much less time for the post-processing step. (3) At the component level, a pre-trained and denser ResNet encoder and random stretching data augmentation can help boost performance for all methods.
For LayoutNet, the post-processing method that works under the equirectangular view performs better. For DuLa-Net and HorizonNet, the post-processing step under ceiling view is more suitable.
%
%
%
We hope our analysis and discoveries can inspire researchers to build up more robust and efficient 3D layout reconstruction methods from a single panoramic image in the future.
%

Our contributions are:
\begin{itemize}
\item We introduce two frameworks, {\layoutnetEX} and {\dulanetEX}, which extend the corresponding state-of-the-art approaches of 3D Manhattan layout reconstruction from an RGB panoramic image. Our approaches compare well in terms of speed and accuracy and achieve the best results for cuboid layout reconstruction.
\item We conduct extensive experiments for {\layoutnetEX}, {\dulanetEX} and another state-of-the-art approach, HorizonNet. We discuss the effects of encoders, post-processing steps, performance for different room shapes, and time consumption. Our investigations can help inspire researchers to build up more robust and efficient approaches for single panoramic image 3D layout reconstruction.
\item We extend the Matterport3D dataset with general Manhattan layout annotations. The annotations contain panoramas depicting room shapes of various complexity. The dataset will be made publicly available. In addition, two depth-based evaluation metrics are introduced for measuring the performance of general Manhattan layout reconstruction.  
\end{itemize}


%% file: fig_illustration.tex
\begin{figure*}
\includegraphics[width=1.0\textwidth]{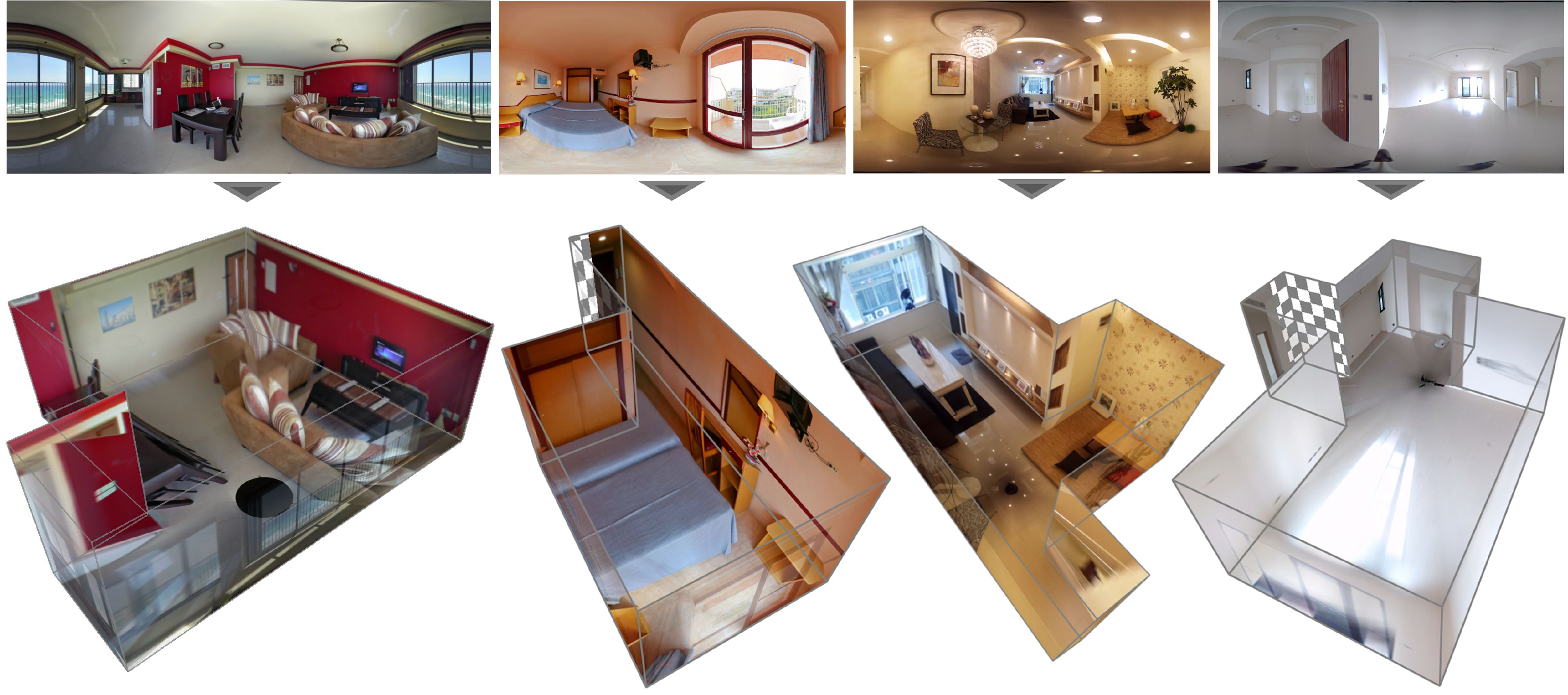}
\caption{Our \textbf{goal} is to reconstruct a 3D manhattan room layouts (bottom) from a single panoramic image (top).}
\label{fig:illustration} 
\end{figure*}

%% file: Related_work.tex
\section{Related Work}
\label{sec:relatedwork}

There are numerous papers that propose solutions for estimating a 3D
room layout from a single image. The solutions differ in the layout shapes~(\ie cuboid layout \vs general Manhattan layout), inputs~(\ie perspective \vs panoramic image), and methods to predict geometric features and fit model parameters. 

In terms of room layout assumptions, a popular choice is the ``Manhattan world" assumption~\cite{coughlan1999manhattan}, meaning that all walls are aligned with a canonical coordinate system~\cite{coughlan1999manhattan,ramalingam2013manhattan}. To make the problem easier, a more restrictive assumption is that the room is a cuboid~\cite{hedau2009recovering, dasgupta2016delay, lee2017roomnet}, \ie there are exactly four room corners in the top-down view.
Recent state-of-the-art methods~\cite{zou2018layoutnet,yang2019dula,sun2019horizonnet} adopt the Manhattan world assumption but allow for room layout with arbitrary complexity. 

In terms of the type of input images, the images may differ in the FoV (field of view) - ranging from being monocular (\ie taken from a standard camera) to 360$^\circ$ panoramas, and whether depth information is provided. The methods are then largely depending on the input image types. It is probably most difficult problem when only a monocular RGB image is given. Typically, geometric (\eg lines and corners)~\cite{lee2009geometric, hedau2009recovering, ramalingam2013lifting} and/or semantic (\eg segmentation into different regions~\cite{hoiem2005geometric, hoiem2007recovering} and volumetric reasoning~\cite{gupta2010estimating}) "cues" are extracted from the input image, a set of room layout hypotheses is generated, and then an optimization or voting process is taken to rank and select the best one among the hypotheses. 

Traditional methods treat the task as an optimization problem. Flint~\etal~\cite{flint2010dynamic} propose a dynamic programming approach for the room layout vectorization. They further apply a graphical model in the context of a moving camera. A more recent work by Delage ~\etal~\cite{delage2006dynamic} fit floor/wall boundaries in a perspective image taken by a level camera to create a 3D model under the Manhattan world assumption using dynamic Bayesian networks. Most methods are based on finding best-fitting hypotheses among detected line segments~\cite{lee2009geometric}, vanishing points~\cite{hedau2009recovering}, or geometric contexts~\cite{hoiem2005geometric}. Subsequent works follow a similar approach, with improvements to layout generation~\cite{schwing2012efficient,schwing2012efficient_eccv,ramalingam2013manhattan}, features for scoring layouts~\cite{schwing2012efficient_eccv,ramalingam2013manhattan}, and incorporation of object hypotheses~\cite{hedau2010thinking,gupta2010estimating,del2012bayesian,del2013understanding,zhao2013scene} or other context.

Recently, neural network-based methods took stride in tackling this problem. There exist methods that train deep network to classify pixels into layout surfaces~(\eg walls, floor, ceiling)~\cite{dasgupta2016delay,izadinia2017im2cad}, boundaries~\cite{mallya2015learning}, corners~\cite{lee2017roomnet}, or a combination~\cite{ren2016coarse}. A trend is that the neural networks generate higher and higher levels of information - starting from line segments~\cite{mallya2015learning, stpio}, surface labels~\cite{dasgupta2016delay}, to room types~\cite{lee2017roomnet} and room boundaries and corners~\cite{zou2018layoutnet}, to faciliate the final layout generation process. Recent methods push the edge further by using neural networks to directly predict a 2D floor plan~\cite{yang2019dula} or as three 1D vectors that concisely encode the room layout~\cite{sun2019horizonnet}. In both cases, the final room layouts are reconstructed by a simple post-processing step.


Another line of works aims to leverage the extra depth information for room model reconstruction, including utilizing single depth image for 3D scene reconstruction~\cite{zhang2013estimating, zou2019complete, liu2016layered}, and scene reconstructions from point clouds~\cite{newcombe2011kinectfusion, monszpart2015rapter, liu2018floornet, cabral2014piecewise}.
Liu~\etal~\cite{liu2015rent3d} present Rent3D, which takes advantage of a known floor plan. Note that neither estimated depths nor reconstructed 3D scenes necessarily equate a clean room layout as such inputs may contain clutters.

\paragraph{360$^\circ$ panorama:}
The seminal work by Zhang~\etal~\cite{zhang2014panocontext} advocates the use of 360$^\circ$ panoramas for indoor scene understanding, for the reason that the FOV of 360$^\circ$ panoramas is much more expansive. Work in this direction flourished, including methods based on optimization approaches over geometric~\cite{Fukano2016RoomRF,pintore2016omnidirectional,yang2016efficient,yang2016efficient,xu2017pano2cad} and/or semantic cues~\cite{xu2017pano2cad,automatic} and later based on neural networks~\cite{lee2017roomnet,zou2018layoutnet}. Most methods rely on leveraging existing techniques for single perspective images on samples taken from the input panorama.
The LayoutNet introduced by Zou~\etal~\cite{zou2018layoutnet} was the first approach to predict room layout directly on panorama, which led to better performance. 
Yang~\etal~\cite{yang2019dula} and Pintore~\etal~\cite{pintore2016omnidirectional} follow the similar idea and propose to predict directly in the top-down view converted from input panorama.
%
%
In this manner, the vertical lines in the panorama become radial lines emanated from the image center. An advantage of this representation is that the room layout becomes a closed loop in 2D that can be extracted more easily. As mentioned in~\cite{yang2019dula}, the ceiling view is arguably better as it provides a clutter-free view of the room layout.

%% file: Dataset.tex
\section{Datasets}
\label{sec:dataset}

Datasets with detailed ground truth 3D room layouts play a crucial role for both network training and performance validation. In this work, we use three public datasets for the evaluation, which are {\em PanoContext}~\cite{zhang2014panocontext}, {\em Stanford 2D-3D}~\cite{stfd2d3d}, and {\em Matterport3D}~\cite{chang2017matterport3d}. All three datasets are composed of RGB(D) panoramic images of various indoor scene types and differ from each other in the following intrinsic properties: (1) the complexity of room layout; (2) the diversity of scene types; and (3) the scale of dataset. For those datasets lack of ground truth 3D layouts, we further extend their annotations with detailed 3D layouts using an interactive annotator, PanoAnnotator~\cite{yang2018panoannotator}. A few sample panoramic images from the chosen datasets are shown in \figref{fig:dataset}. We will briefly describe each dataset and discuss differences as follows.

\subsection{PanoContext Dataset}

\input{fig_dataset.tex}


PanoContext~\cite{zhang2014panocontext} dataset contains $514$ RGB panoramic images of two indoor environments, \ie bedrooms and living rooms, and all the images are annotated as cuboid layouts. For the evaluation, we follow the official train-test split and further carefully split 10\% validation images from the training samples such that similar rooms do not appear in the training split.


\subsection{Stanford 2D-3D Dataset}
Stanford 2D-3D~\cite{stfd2d3d} dataset contains $552$ RGB panoramic images collected from $6$ large-scale indoor environments, including offices, classrooms, and other open spaces like corridors. Since the original dataset does not provide ground truth layout annotations, we manually labeled the cuboid layouts using the PanoAnnotator. The Stanford 2D-3D dataset is more challenging than PanoContext as the images have smaller vertical FOV and more occlusions on the wall-floor boundaries. We follow the official train-val-test split for evaluation.


\subsection{Our Labeled MatterportLayout Dataset}
We carefully selected $2295$ RGBD panoramic images from Matterport3D~\cite{chang2017matterport3d} dataset and extended the annotations with ground truth 3D layouts. We call our collected and relabeled subset the {\em MatterportLayout} dataset.

Matterport3D~\cite{chang2017matterport3d} dataset is a large-scale RGB-D dataset containing over ten thousand RGB-D panoramic images collected from 90 building-scale scenes. Matterport3D has the following advantages over the other datasets:
\begin{enumerate}
    \item covers a larger variety of room layouts (\eg cuboid, ``L''-shape, ``T''-shape rooms, etc) and over 12 indoor environments (\eg bedroom, office, bathroom and hallway, etc);
    \item has aligned ground truth depth for each image, allowing quantitative evaluations for layout depth estimation; and
    \item is three times larger in scale than PanoContext and Stanford 2D-3D, providing rich data for training and evaluating our approaches.
\end{enumerate}


Note that there also exists the Realtor360 dataset introduced in~\cite{yang2019dula}, which contains over $2500$ indoor panoramas and annotated 3D room layouts. However, Realtor360 currently could not be made publicly available due to legal privacy issues. 
\input{fig_dataset_barchart}
Moreover, as shown in~\figref{fig:dataset_corner_compare}, our MatterportLayout dataset covers a more diverse range of layout types than Realtor360 .
%
%
%
%
The detailed annotation procedure and dataset statistics are elaborated as follows. 


%% file: fig_dataset.tex
\begin{figure}[!t]
\includegraphics[width=0.49\textwidth]{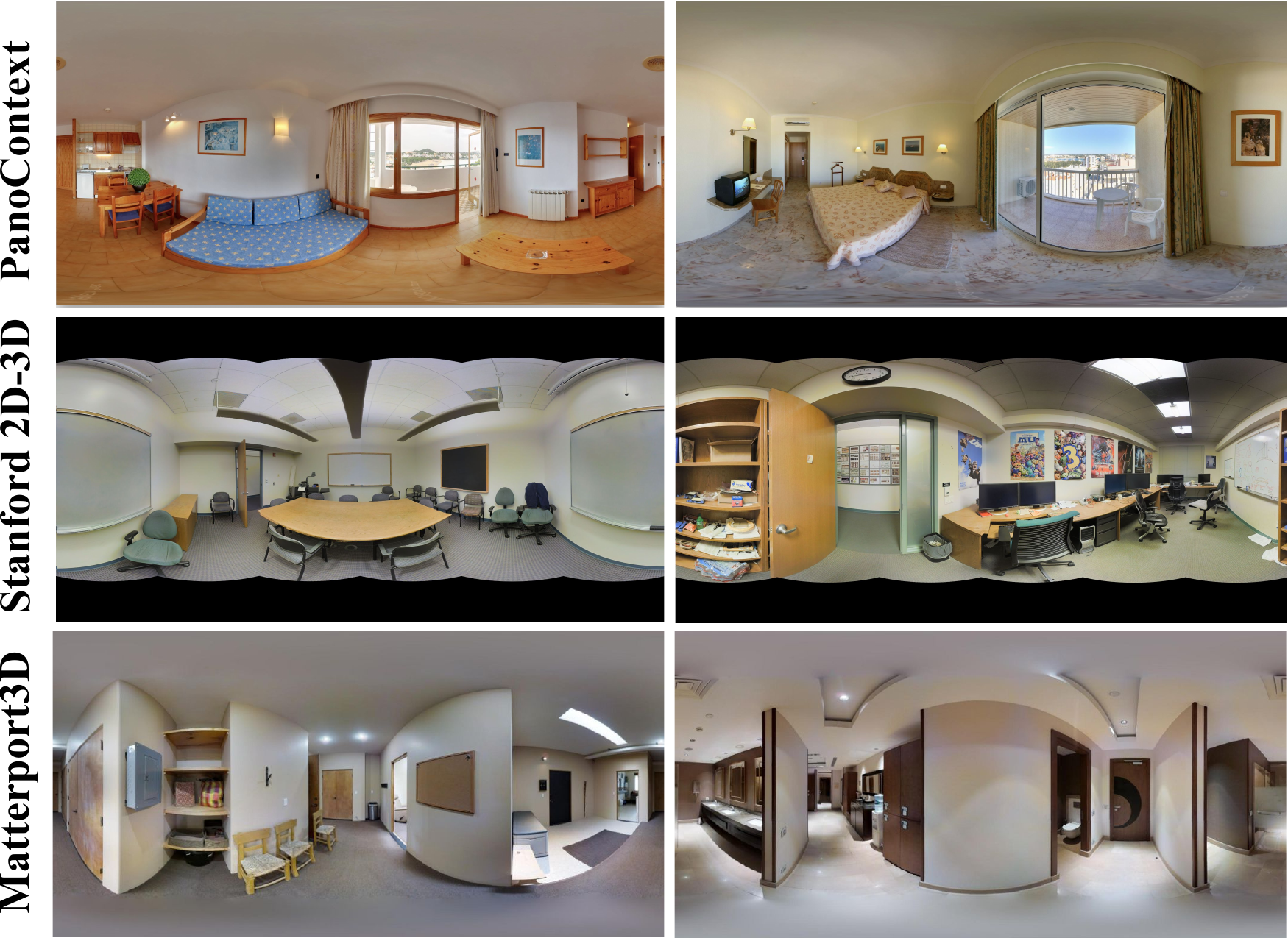}
\caption{Sample images from the three datasets we use for evaluation: PanoContext dataset~(top), Stanford 2D-3D dataset~(middle) and our labeled MatterportLayout dataset~(bottom).}
\label{fig:dataset} 
\end{figure}

%% file: fig_dataset_barchart.tex
\begin{figure*}[!t]
\includegraphics[width=\textwidth]{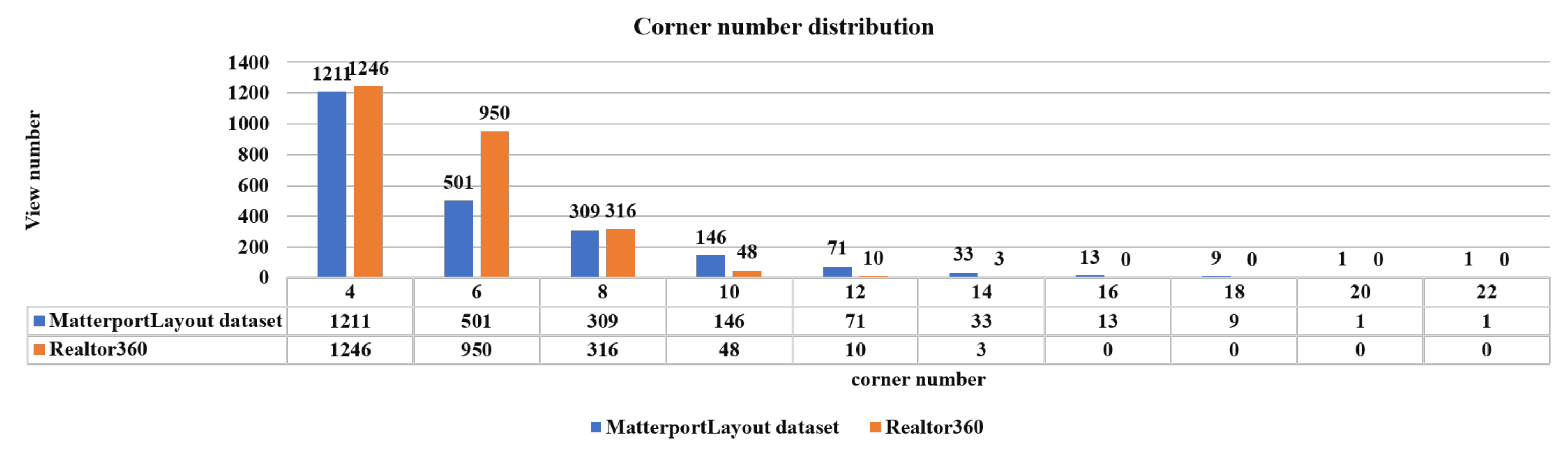}
\caption{The comparison of the layout type distribution of MatterportLayout and Realter360 datasets. We categorize each panoramic image into different types based on the number of layout corners in the ceiling view. Our MatterportLayout dataset covers a more diverse range of indoor layout types.
}
\label{fig:dataset_corner_compare} 
\end{figure*}

%% file: Matterport3D.tex
\input{fig_panannotator}
\input{fig_matterport3D.tex}

\subsubsection{Annotation Process of~\datasetName} 
We first exclude images in the Matterport3D dataset with (1) open 3D space like corridors and stairs, (2) non-flat floors and non-Manhattan 3D layouts, (3) outdoors scenes and (4) artifacts resulting from stitching perspective views.


We use PanoAnnotator~\cite{yang2018panoannotator} to annotate ground truth 3D layouts. The interface of the annotation tool is shown in \figref{fig:panoannotator}. The annotation pipeline involves four steps: (1) Select one of the panoramas in the dataset; (2) load an initial layout (this step is optional);
%
%
(3) add or remove the 3D wall pieces to match the number of walls that the given layout has; (4) push or pull all the walls to match the panorama; (5) push the ceiling and floor to match the panorama. Finally, we obtain a dataset of $2295$ RGB-D panoramas with detailed layout annotations. 
For all the layouts in the dataset, we started with initial layouts from~\cite{fuenmatterport}, which is also annotated with PanoAnnotator.
We refined each layout to achieve accurate room corners, ceiling height, and walls position. More than 25\% of initial layouts were adjusted significantly with at least 10\% difference in 3D IoU. 

%
  
%
To evaluate the quality of estimated 3D layouts using the depth measurements, we further process the aligned ground truth depth maps to remove pixels that belong to foreground objects (\eg furniture). Specifically, we align the ground truth depth map to the rendered depth map of the annotated 3D layout and mask out inconsistent pixels between two depth maps. For the alignment, we scale the rendered depth by normalizing camera height to 1.6m. We then mask out pixels in the ground truth depth map that are more than 0.15m away from their counterparts in the rendered depth map. In our experiment, we use the unmasked pixels for evaluation. See~\figref{fig:matterport3d} for some examples.

\input{tbl_matterport3D.tex}


\subsubsection{\datasetName~Dataset Statistics} 
We use approximately 70\%, 10\%, and 20\% of the data for training, validation, and testing. Images from the same room do not appear in different sets. Moreover, we ensure that the proportions of rooms with the same 3D shape in each set are similar. We show in Table~\ref{tab:matterport3d} the total numbers of images annotated for different 3D room shapes. The 3D room shape is classified according to the numbers of corners in ceiling view: cuboid shape has four corners,``L''-shape room has six corners, ``T''-shape has eight corners, etc. The {\datasetName} dataset covers a large variety of 3D room shapes, with approximately 52\% cuboid rooms, 22\% ``L''-shape rooms, 13\% ``T''-shape rooms, and 13\% more complex room shapes. The train, validation, and test sets have similar distributions of different room shapes, making our experiments reliable for both training and testing.

%% file: fig_panannotator.tex
\begin{figure}[!t]
\includegraphics[width=0.49\textwidth]{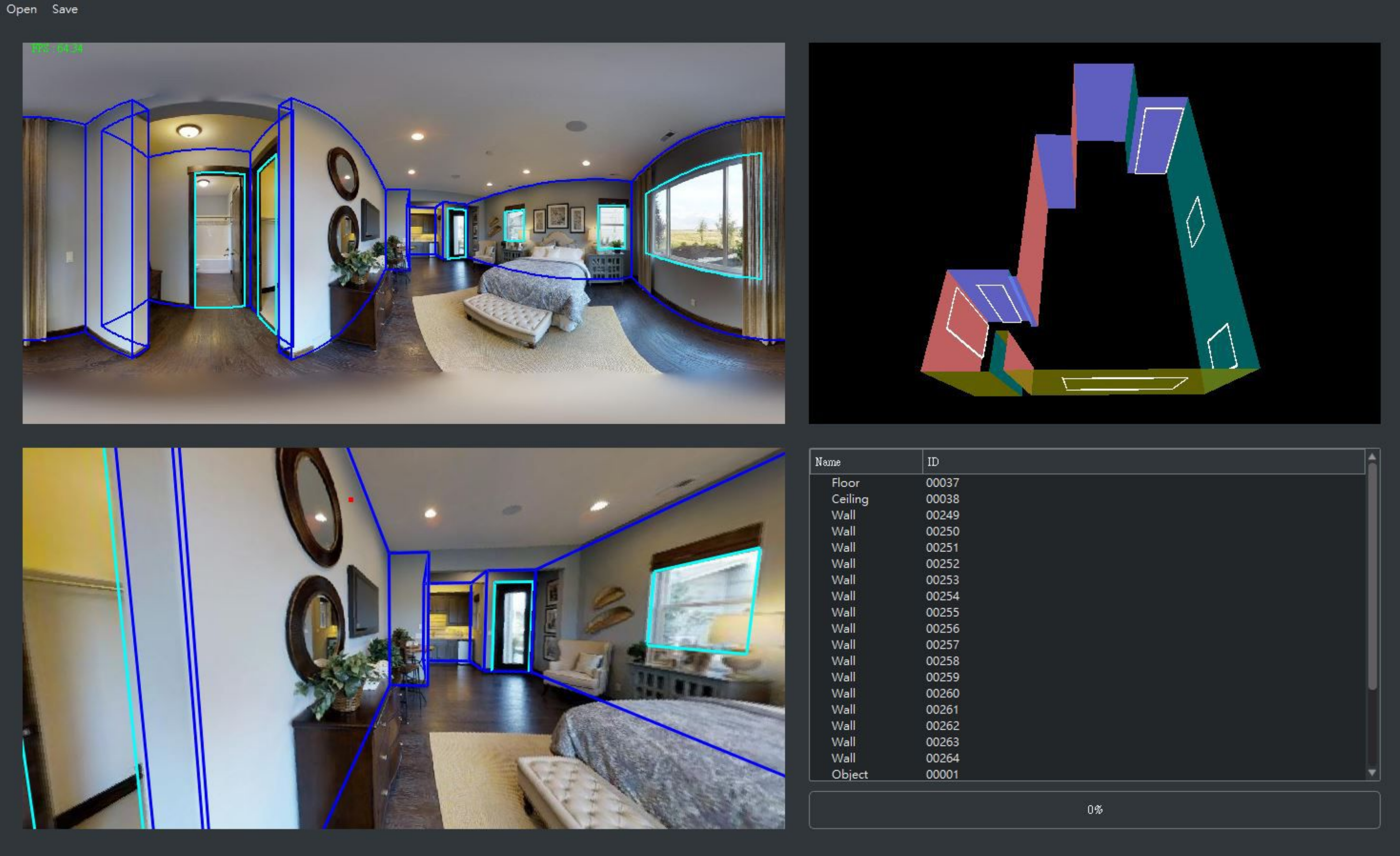}
\caption{We use PanoAnnotator~\cite{yang2018panoannotator} to annotate our MatterportLayout dataset. This figure illustrates the interface of the tool with PanoView (top-left), ResultView (top-right), MonoView (bottom-left) and ListView (bottom-right). 
}
\label{fig:panoannotator}
\end{figure}

%% file: fig_matterport3D.tex
\begin{figure*}[!t]
\includegraphics[width=1.0\textwidth]{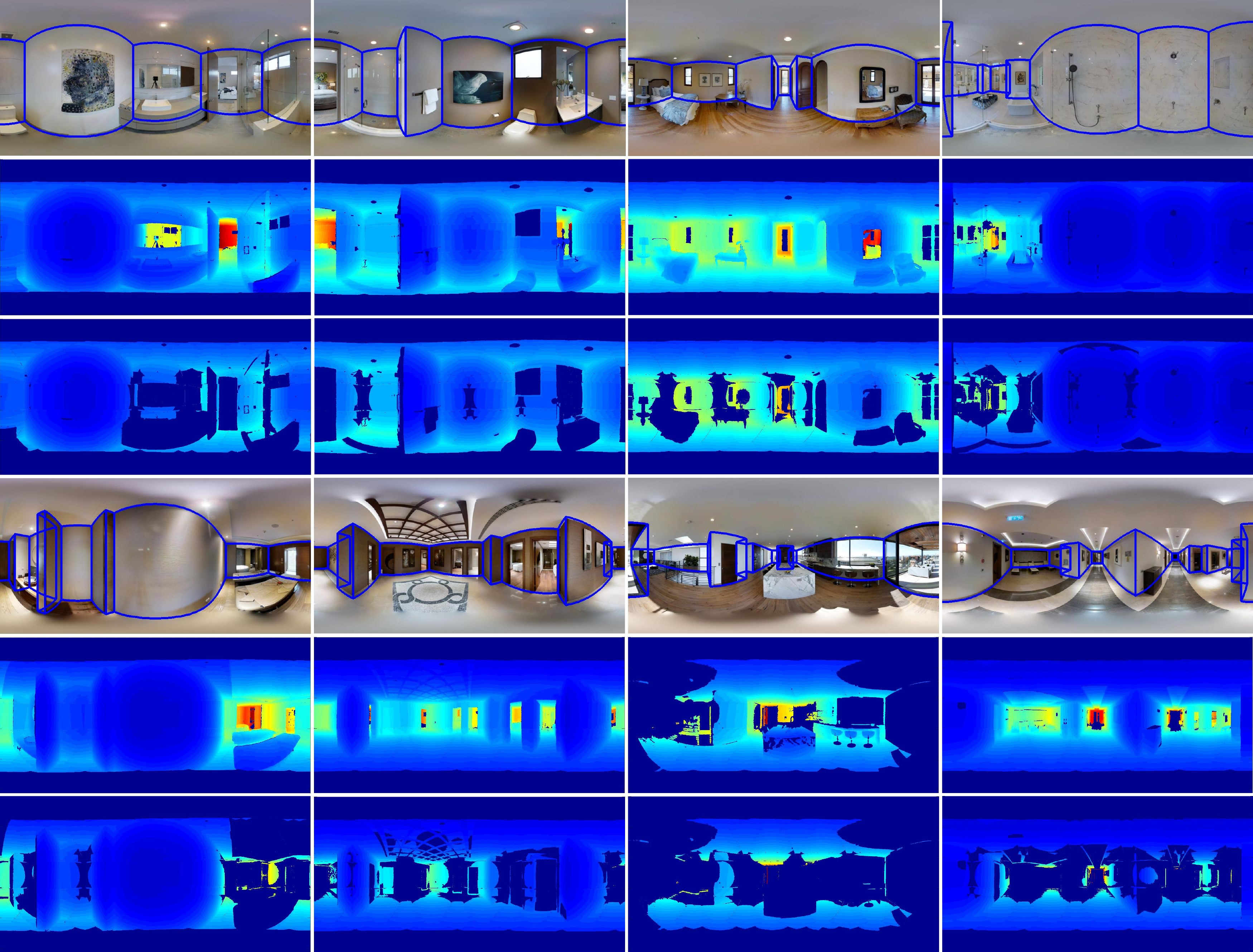}
\caption{Sample images of {\datasetName} dataset. Each image has different room corners in ceiling view, from 4 corners~(cuboid, top left), 6 corners~(``L''-shape, 1st row, 2nd column), to 18 corners~(bottom right). We show in each sample: the ground truth Manhattan 3D layout labeled on the RGB panorama under equirectangular view, the original depth map and the processed depth map where the occluded regions are masked out.}
\label{fig:matterport3d} 
\end{figure*}

%% file: tbl_matterport3D.tex
\begin{table}[!b]
\caption{The {\datasetName} dataset statistics. Total number of images for different 3D room shapes categorized by number of corners in the ceiling view.}
\begin{center}
\label{tab:matterport3d}
\resizebox{0.5\textwidth}{!}{
\begin{tabular}{ccccccccccc}
\hline
\# of Corners & 4 & 6 & 8 & 10 & 12 & 14 & 16 & 18 & 20 & 22\\
\hline\hline
Full dataet & 1211 & 501 & 309& 146 & 71 & 33 & 13 & 9 &1 &1\\
\hline
Train set &  841 & 371 & 225 & 114 & 51 & 27 & 9 & 7 & 1 & 1\\
Validation Set & 108 & 46 & 21 & 7 & 5 & 1 & 2 & 0 & 0 & 0\\
Test set & 262 & 84 & 63 & 25 & 15 & 5 & 2 & 2 & 0 & 0\\
\hline
\end{tabular}
}
\end{center}
\end{table}

%% file: Method_overview.tex
\input{tbl_taxonomy_1.tex}

\section{Methods Overview}
\label{sec:network}
In this section, we introduce the common framework, the variants, and the impact of the design decisions of recently proposed approaches for 3D Manhattan layout reconstruction from a single panoramic image. \tabref{tab:taxonomy_1} summarizes the key design choices that LayoutNet~(\figref{fig:layoutnet-overview}), DuLa-Net~(\figref{fig:dulanet-overview}) and HorizonNet~(\figref{fig:horizonnet-overview}) originally proposed in their papers respectively. Though all three methods follow the same general framework, they differ in the details. We unify some of the designs and training details and propose our modified LayoutNet and DuLa-Net methods as follows, which show better performance compared with the original ones.


\subsection{General framework}\label{subsec:method}
The general framework can be decomposed into three parts. First, we discuss in~\secref{subsec:preprocess} the input and pre-processing step. Second, we introduce the network design of encoder in~\secref{subsec:encoder}, the decoder of layout pixel predictions in~\secref{subsec:decoder}, and the training loss for each method in~\secref{subsec:loss}. Finally, we discuss the structured layout fitting in~\secref{subsec:opt}.

\subsection{Input and Pre-processing}
\label{subsec:preprocess}
Given the input as a panorama that covers a $360^\circ$ horizontal field of view, the first step of all methods is to align the image to have a horizontal floor plane. The alignment, first proposed by LayoutNet and then inherited by DuLa-Net and HorizonNet, ensures that wall-wall boundaries are vertical lines and substantially reduces error. The alignment goes as follows. First, we estimate the orientation of floor plan under spherical projection using Zhang~\etal's approach~\cite{zhang2014panocontext} (i.e., selecting long line segments using the Line Segment Detector~(LSD)~\cite{von2008lsd} in each overlapping perspective view), then we vote for three mutually orthogonal vanishing directions using the Hough Transform. Afterward, we rotate the scene and re-project it to the 2D equirectangular projection. 

The aligned panoramic image is used for all three methods as input. For better predicting layout elements, LayoutNet and DuLa-Net utilize additional inputs as follows.

LayoutNet additionally concatenates a $512\times1024$ Manhattan line feature map lying on three orthogonal vanishing directions using the alignment method as described in the previous paragraph. The Manhattan line feature map provides additional input features that were shown to have improved the performance quantitatively~\cite{zou2018layoutnet}. 

To feed its two-branch network, DuLa-Net creates another ceiling-view perspective image projected from the input panorama image using an E2P module described in the following.


For every pixel in the perspective image (assumed to be square with dimension $w$ squared) at position $(p_x,p_y)$, the position of the corresponding pixel in the equirectangular panorama, $(p\textprime_x,p\textprime_y)$, $-1 \le p\textprime_x \le 1, -1 \le p\textprime_y \le 1$, is derived as follows. First, the field of view of the pinhole camera of the perspective image is defined as $FoV$. Then, the focal length can be derived as:
$$
f = 0.5 * w * \cot(0.5 * \mathrm{FoV})~,
$$
$(p_x, p_y, f)$ is the 3D position of the pixel in the perspective image in the camera space. It is then rotated by 90$^\circ$ or -90$^\circ$ along the x-axis~(counter-clockwise) if the camera is looking upward~(\eg looking at the ceiling) or downward~(\eg looking at the floor), respectively. Next, the rotated 3D position is projected to the equirectangular space. To do so, the rotated 3D position is first projected onto a unit sphere by vector normalization, the resulting 3D position on the unit sphere is denoted as $(s_x, s_y, s_z)$, and then the following formula is applied to project $(s_x, s_y, s_z)$ back to $(p\textprime_x,p\textprime_y)$, which is the corresponding 2D position in the equirectangular panorama:
\begin{align}
(p\textprime_{x}, p\textprime_{y}) &= (\frac{arctan_2 (\frac{s_{x}}{s_{z}})}{\pi}~, \frac{arcsin (s_{y})}{0.5\pi}),
\end{align}
Finally, $(p\textprime_x,p\textprime_y)$ is used to interpolate a pixel value from the panorama. Note that this process is differentiable so it can be used in conjunction with back-propagation. DuLa-Net peforms E2P with a $FoV$ of $160^\circ$ and produces a perspective image of $512\times512$.

The need for ceiling view branch in DuLaNet and E2P module can be explained by two reasons: (1) DuLa-Net is able to extract more features from ceiling view and (2) referring to the Table 2 in the original paper~\cite{yang2019dula}, the best performance comes from two branches with E2P module.

\input{fig_layoutnet_overview.tex}
\input{fig_dulanet_overview.tex}

\subsection{Encoder}
\label{subsec:encoder}
%


Regarding the network designs, the original LayoutNet uses SegNet as the encoder while DuLa-Net and HorizonNet use ResNet. For our modified version, we use ResNet uniformly because we find that it shows better performance in capturing layout features than SegNet in experiments~(\secref{text:abla}, \secref{tab:cuboid_abla}). Details are as follows.

The ResNet encoder receives a $512\times1024$ RGB panoramic image under equirectangular view as input.  For the three methods, there are differences in the last encoding layers based on their different designs of decoders for different output spaces: for LayoutNet, the last fully connected layer and the average pooling layer of the ResNet encoder are removed. For DuLa-Net, it uses a separate ResNet encoder for both {\panoBranch} and {\ceilBranch}. The {\panoBranch} $B_P$ has an output dimension of $16 \times 32 \times 512$. For the {\ceilBranch} $B_C$, the output dimension is $16 \times 16 \times 512$. Finally, for HorizonNet, 
it performs a separate convolution for each of the feature maps produced by each block of the ResNet encoder. The convolution down samples each map by 8 in height and 16 in width, with feature size up-sampled to 256. The feature maps are then reshaped to 256 x 1 x 256 and concatenated based on the first dimension, producing the final bottleneck feature.

\subsection{Layout Prediction}\label{subsec:decoder}
Next, we discuss the decoders. In general, the decoders output predictions of layout pixels in the form of either corner and boundary positions under equirectangular view, or semantic floor plan map under ceiling view. We describe each type of prediction as follows.

\subsubsection{Equirectangular-view prediction}

Decoders of LayoutNet and HorizonNet output layout predictions in equirectangular view solely, while DuLa-Net's two-branch network design means that decoders output predictions in both equirectangular view and ceiling view.

Both LayoutNet and HorizonNet predict layout corners and boundaries under equirectangular projections. For LayoutNet, the decoder consists of two branches. The top branch, the layout boundary map~($\boldsymbol{m_E}$) predictor, decodes the bottleneck feature into a 2D feature map with the same resolution as the input. $\boldsymbol{m_E}$ is a 3-channel probability prediction of wall-wall, ceiling-wall and wall-floor boundary on the panorama, for both visible and occluded boundaries. The boundary predictor contains $7$ layers of nearest neighbor up-sampling operation, each followed by a convolution layer with kernel size of $3\times 3$, and the feature size is halved through layers from $2048$. The final layer is a Sigmoid operation. Skip connections are added to each convolution layer following the spirit of the U-Net structure~\cite{ronneberger2015u}, in order to prevent shifting of prediction results from the up-sampling step. The lower branch, the 2D layout corner map~($\boldsymbol{m_C}$) predictor, follows the same structure as the boundary map predictor and additionally receives skip connections from the top branch for each convolution layer. This stems from the intuition that layout boundaries imply corner positions, especially for the case when a corner is occluded. It's shown in~\cite{zou2018layoutnet} that the joint prediction helps improve the accuracy of the both maps, leading to a better 3D reconstruction result. 
We exclude the 3D regressor proposed in~\cite{zou2018layoutnet} as the regressor is shown to be ineffective in the original paper.

The output space of LayoutNet is $O(HW)$, where $H$ and $W$ are the height and width of the input image. This dense prediction limits the network to use deeper encoders such as ResNet-50 or complex decoders to improve performance further. HorizonNet simplifies LayoutNet's prediction by predicting three 1-D vectors with 1024 dimensions instead of two 512x1024 probability maps. The three vectors represent the ceiling-wall and the floor-wall boundary position, and the existence of wall-wall boundary~(or corner) of each image column. HorizonNet further applies an RNN block to refine the vector predictions, which considerably help boost performance as reported in~\cite{sun2019horizonnet}.

For DuLa-Net, its {\panoBranch} $B_P$ predicts floor-ceiling probability map $M_{FC}$ under equirectangular view. $M_{FC}$ has the same resolution as the input. A pixels in $M_{FC}$ with higher value means a higher probability to be ceiling or floor. The decoder of $B_P$ consists of 6 layers. Each of the first 5 layers contains the nearest neighbor up-sampling operation followed by a $3\times3$ convolution layer and ReLU activation function, the channel number is halved from 512~(if using ResNet18 as an encoder). The final layer of the decoder replaces the ReLU by Sigmoid to ensure the data range is in $[0, 1]$. The second branch of DuLa-Net predicts 2D probability map under ceiling view which will be introduced in the next paragraph.

\subsubsection{Ceiling-view prediction}

Ceiling-view prediction is exclusive to DuLa-Net. Contrary to its {\panoBranch} decoder, its {\ceilBranch} decoder, $B_C$, outputs a $512\times512$ probability map in the ceiling view. A pixel with higher value indicates a higher probability to be part of the ceiling. Decoders in both branches have the same architecture.
DuLa-Net then fuses the feature map from the {\panoBranch} to the {\ceilBranch} through the E2P projection module as described in~\secref{subsec:encoder}.
Applying fusion techniques 
increases the prediction accuracy. It is conjectured that, in a ceiling-view image, the areas near the image boundary~(where some useful visual clues such as shadows and furniture arrangements exist) are more distorted, which can have a detrimental effect for the {\ceilBranch} to infer room structures. By fusing features from the {\panoBranch}~(in which distortion is less severe), performance of the {\ceilBranch} can be improved.

DuLa-Net applies fusions before each of the first five layers of the decoders. For each fusion connection, an E2P conversion with the $FoV$ set to 160$^\circ$ is taken to project the features under the equirectangular view to the perspective ceiling view. Each fusion works as follows:
\begin{align}
\label{equ:fusion}
    f_{B_C}^* = f_{B_C} + \frac{\alpha}{\beta^i} \times f_{B_P},~i \in \{0, 1, 2, 3, 4\},
\end{align}
where $f_{B_C}$ is the feature from {\ceilBranch} and $f_{B_P}$ is the feature from {\panoBranch} after applying the {\EtoP} conversion. $\alpha$ and $\beta$ are the decay coefficients. $i$ is the index of the layer. After each fusion, the merged feature, $f_{B_C}^*$, is sent into the next layer of ceiling-view decoder. 

Note that DuLa-Net's 2D floor plan prediction cannot predict 3D layout height, which is an important parameter for 3D layout reconstruction. To infer the layout height, three fully connected layers are added to the middlemost feature of {\panoBranch}. The dimensions of the three layers are 256, 64, and 1. To make the regression of the layout height more robust, dropout layers are added after the first two layers. To take the middlemost feature as input, DuLa-Net first applies average along channel dimensions, which produces a 1-D feature with 512 dimensions, and take it as the input of the fully connected layers.

\subsection{Loss Function}
\label{subsec:loss}
We discuss the loss functions for each of the original methods.

\subsubsection{LayoutNet}

The overall loss function is:
\begin{align}\label{equ:loss_layoutnet}
    L(\boldsymbol{m_E},\boldsymbol{m_C},\boldsymbol{d}) &= -\alpha \frac{1}{n}\sum_{p\in \boldsymbol{m_E}}\big(p^{*}\log p +(1-p^{*})\log (1-p) \big)\nonumber\\
    &-\beta \frac{1}{n}\sum_{q\in \boldsymbol{m_C}}\big(q^{*}\log q +(1-q^{*})\log (1-q) \big)
\end{align}
Here $\boldsymbol{m_E}$ is the probability that each image pixel is on the boundary between two walls; $\boldsymbol{m_C}$ is the probability that each image pixel is on a corner; $p$ and $q$ are pixel probabilities of edge and corner with ground truth values of $\hat{p}$ and $\hat{q}$, respectively.  The loss is the summation over the binary cross entropy error of the predicted pixel probability in $\boldsymbol{m_E}$ and $\boldsymbol{m_C}$ compared with ground truth.

\subsubsection{DuLa-Net}

The overall loss function is:
\begin{align}
    \label{equ:loss_dulanet}
    &L = E_b(M_{FC}, M_{FC}^*) + E_b(M_{FP}, M_{FP}^*) + \gamma E_{L1}(H, H^*),
\end{align}
Here for $M_{FC}$ and $M_{FP}$, we apply binary cross entropy loss:
\begin{align}
    E_b(x, x^*) = -\sum_{i} {x}_i^*\log({x}_i) + (1-{x}_i^*)\log(1-{x}_i).
\end{align}
For $H$ (layout height), we use L1-loss:
\begin{align}
    E_{L1}(x, x^*) = \sum_{i} |x_i - x^*_i|.
\end{align}
where $M_{FC}^*$, $M_{FP}^*$ and $H^*$ are the ground truth of $M_{FC}$, $M_{FP}$, and $H$.

\subsubsection{HorizonNet}

For the three channel 1-D prediction, HorizonNet applies L1-Loss for regressing the ceiling-wall boundary and floor-wall boundary position, and uses binary cross entropy loss for the wall-wall corner existence prediction.

\subsection{Structured Layout Fitting}
\label{subsec:opt}
Given the 2D predictions (i.e., corners, boundaries and ceiling-view floor plans), the camera position and 3D layout can be directly recovered, up to a scale and translation, by assuming that bottom corners are on the same ground plane and that the top corners are directly above the bottom ones. The layout shape is constrained to be Manhattan, so that intersecting walls are perpendicular, \eg like a cuboid or ``L"-shape in a ceiling view. The final output is a sparse and compact planar 3D Manhattan layout. The optimization can be performed under the equirectangular view or the ceiling view. The former approach is taken by LayoutNet while the latter is taken by DuLa-Net and HorizonNet. In the following, we explain our modified version of the LayoutNet method and the original DuLa-Net and HorizonNet methods in details.

\subsubsection{Equirectangular-view fitting}

Since LayoutNet's network outputs (i.e., 2D corner and boundary probability maps) are under the equirectangular view, the 3D layout parameters are optimized to fit the predicted 2D maps. The initial 2D corner predictions are obtained from the corner probability map (the output of the network) as follows. First, the responses are summed across rows, to get a summed response for each column.  Then, local maxima are found in the column responses, with distance between local maxima of at least 20 pixels.  Finally, the two largest peaks are found along the selected columns. These 2D corners might not satisfy Manhattan constraints, so we perform optimization to refine the estimates. 

The ceiling level is initialized as the average~(mean) of 3D upper-corner heights, and then optimize for a better fitting room layout, relying on both corner and boundary information to evaluate 3D layout candidate $L$:
\begin{align}
Score(L)&= w_{junc}\frac{1}{|C|}\sum_{l_c\in C} P_{\text{corner}}(l_c)\nonumber\\
&+ w_{ceil}\frac{1}{|L_e|}\sum_{l_e\in L_e} P_{\text{ceil}}(l_e)\nonumber\\
&+ w_{floor}\frac{1}{|L_f|}\sum_{l_f\in L_f} P_{\text{floor}}(l_f)
\label{equ:layotnetopt}
\end{align}
where $C$ denotes the 2D projected corner positions of $L$. Cardinality of $L$ is \#walls$\times$ 2. The nearby corners are connected on the image to obtain $L_e$ which is the set of projected wall-ceiling boundaries, and $L_f$ which is the set of projected wall-floor boundaries~(each with cardinality of \#walls). $P_{\text{corner}}(\cdot)$ denotes the pixel-wise probability value on the predicted $\boldsymbol{m_C}$. $P_{\text{ceil}}(\cdot)$ and $P_{\text{floor}}(\cdot)$ denote the probability on $\boldsymbol{m_E}$. LayoutNet finds that adding wall-wall boundaries in the scoring function helps less, since the vertical pairs of predicted corners already reveals the wall-wall boundaries information. 

Note that the cost function in~\eqnref{equ:layotnetopt} is slightly different from the cost function originally proposed in LayoutNet - we revise the cost function to compute the average response across layout lines instead of the maximum response. In this way, we are able to produce a relatively smoothed space for the gradient ascent based optimization as introduced below.
The originally proposed LayoutNet uses sampling to find the best ranked layout based on the cost function, which is time consuming and is constrained to the pre-defined sampling space. We instead use stochastic gradient ascent~\cite{robbins1951stochastic} to search for local optimum of the cost function~\footnote{We revised the SGD based optimization implemented by Sun~(with different loss term weights): https://github.com/sunset1995/pytorch-layoutnet}. We demonstrate the performance boost by using gradient ascent in experiments~(\secref{text:abla})

Finally, we made a few extensions. As LayoutNet's network prediction might miss occluded corners, which are important for the post-processing step that relies on Manhattan assumption, we adopt HorizonNet's post-processing step to find occluded corners for initialization before performing the fitting refinement in the equirectangular view.

\input{fig_floorplan_fitting.tex}

\subsubsection{Ceiling-view fitting}

DuLa-Net's network outputs 2D floor plan predictions under ceiling view. Given the probability maps ($\fcmap$ and $\fpmap$) and the layout height ($H$) predicted by the network, DuLa-Net reconstructs the final 3D layout in the following two steps:
\begin{enumerate}
    \item Estimating a 2D Manhattan floor plan shape using the probability maps.
    \item Extruding the floor plan shape along its normal according to the layout height.
\end{enumerate}
%
%
For step 1, two intermediate maps, denoted as $\fcmapCeil$ and $\fcmapFloor$, are derived from ceiling pixels and floor pixels of the {\fcName} using the {\EtoP} conversion.
DuLa-Net further uses a scaling factor, $1.6/(H - 1.6)$, to register the $\fcmapFloor$ with $\fcmapCeil$, where the constant $1.6$ is the distance between the camera and the ceiling.

Finally, a {\ffpName} is computed as follows:
\begin{align}
\displaystyle
\fpmap^{fuse} = 0.5*\fpmap + 0.25*\fcmapCeil + 0.25*\fcmapFloor.
\end{align}
\figref{fig:floorplan_fitting} (a) illustrates the above process.
The probability map $\fpmap^{fuse}$ is binarized using a threshold of $0.5$. A bounding rectangle of the largest connected component is computed for later use.
Next, the binary image is converted to a densely sampled piece-wise linear closed loop and simplify it using the Douglas-Peucker algorithm (see~\figref{fig:floorplan_fitting} (b)).
A regression analysis is run on the edges. The edges are clustered into sets of axis-aligned horizontal and vertical lines. 
These lines divide the bounding rectangle into several disjoint grid cells (see~\figref{fig:floorplan_fitting} (c)).
The shape of the 2D floor plan is defined as the union of grid cells where the ratio of floor plan area is greater than $0.5$ (see~\figref{fig:floorplan_fitting} (d)). Note that this post-processing step does not have an implicit constraints on layout shapes~(cuboid or non-cuboid). To evaluate on cuboid room layout, we directly use the bounding rectangle of the largest connected component as the predicted 2D floor plan for DuLa-Net.

For HorizonNet, although the prediction is done under an equirectangular view, the post-processing step is done under a ceiling view. We observe that computing on the ceiling view helps enforcing the constraints that neighboring walls are orthogonal to each other, and to recover occluded wall corners that cannot be detected from equirectangular view. 
First, the layout height is estimated by averaging over the predicted floor and ceiling positions in each column. Second, the scaled ceiling boundary and floor boundary are projected to the ceiling view, same as Dula-Net. Following LayoutNet's approach, HorizonNet then initializes the corner positions by finding the most prominent wall-wall corner points and project them to ceiling view. The orientations of walls are retrieved by computing the first PCA component along the projected lines between two nearby corners. 
%
The projected ceiling boundary is represented by multiple groups of 2D pixel points separated by the wall-wall boundary. It then gives a higher score to the PCA vector line with more 2D pixel points within 0.16 meters and selects the vector that obtains the highest score as the wall in every group.
Finally, the 3D layout is reconstructed.

\subsubsection{Handling occlusions from layout boundaries}

For non-cuboid Manhattan layouts, some of the walls can be occluded from the camera position. LayoutNet finds the best-fit layout shape based on the 2D predictions, which might not be able to recover the occluded layout corners and boundaries. DuLa-Net fits a polygon to the predicted 2D floor plan, which explicitly enforces the neighboring walls to be orthogonal to each other. HorizonNet detects occlusions from layout boundaries by checking the orientation of the first PCA component for nearby layout walls. If two neighboring walls are parallel to each other, HorizonNet will hallucinate the occluded walls. We conjecture that the difference in handling occlusions from layout boundaries is the main reason why LayoutNet performs better than DuLa-Net and HorizonNet for cuboid layouts (no occlusions from layout boundaries) while performing slightly worse for non-cuboid layouts.

\subsection{Implementation Details}
\label{subsec:implementation}
We implement LayoutNet and DuLa-Net using PyTorch. For HorizonNet, we directly use their PyTorch source code available online for comparison. For implementation details, we summarize the data augmentation methods in~\secref{subsec:augmentation} and the training scheme and hyper-parameters in~\secref{exp:implementation}. 

\subsection{Data augmentation}
\label{subsec:augmentation}
We show in~\tabref{tab:taxonomy_2} the summary of the different data augmentations originally proposed in each method. All three methods use horizontal rotation, left-right flipping and luminance change  
to augment the training samples. 
We unify the data augmentation by adding random stretching~(introduced below) to our modified LayoutNet and DuLa-Net methods. 

\subsubsection{Random Stretching}

Random stretching is introduced by HorizonNet. The augmentation utilizes the $360^{\circ}$ property of panoramic images, projects the pixels into 3D space, stretches pixels along 3D axes, re-projects and interpolates pixels to the equirectangular image to augment training data. 
The effectiveness of this approach has been demonstrated in~\cite{sun2019horizonnet}.

\subsubsection{Ground Truth Smoothing}

For LayoutNet, the target 2D boundary and corner maps are both binary maps that consist of thin curves (boundary map) or points (corner map) on the images, respectively. This makes training more difficult. For example, if the network predicts the corner position slightly off the ground truth, a huge penalty will be incurred. Instead, LayoutNet dilates the ground truth boundary and corner map with a factor of 3 and then smooth the image with a Gaussian kernel of $\sigma = 20$. 
Note that even after smoothing, the target image still contains $\sim 95\%$ zero values, so the back propagated gradients of the background pixels is re-weighted by multiplying with $0.2$. 

This strategy is also taken by HorizonNet for its wall-wall corner existence prediction. It is not taken by DuLa-Net since it predicts the complete floor plan map with clear boundaries.

\input{tbl_taxonomy_2.tex}

\subsection{Training Scheme and Parameters}
\label{exp:implementation}
For our modified LayoutNet and DuLa-Net methods, we use pre-trained weights on ImageNet to initialize the ResNet encoders. We perform random stretching with stretching factors $k_x=1$ and $k_z=2$. For each method, we use the same hyper-parameters for evaluating on the different datasets.

\input{tbl_cuboid_pano.tex}
\input{tbl_cuboid_stdn.tex}

\subsubsection{LayoutNet training}

LayoutNet uses the ADAM~\cite{kingma2014adam} optimizer with $\beta_1=0.9$ and $\beta_2=0.999$ to update network parameters. The network learning rate is $1e^{-4}$. 
To train the network, we first train the layout boundary prediction branch, then fix the weights of boundary branch and train the corner prediction branch, and finally we train the whole network end-to-end. 
To avoid the unstable learning of the batch normalization layer in ResNet encoder due to smaller batch size, we freeze the parameters of the batch normalization~(bn) layer when training end-to-end. The batch size for ResNet-18 and ResNet-34 encoder is 4, while the batch size for ResNet-50 is 2~(Which is too small to have a stable training of the bn layer, leading performance drops comparing with LayoutNet using ResNet-18 or ResNet-34 encoder as shown in~\tabref{tab:cuboid_pano} and~\tabref{tab:cuboid_stdn} in experiments). We set the term weights in~\eqnref{equ:loss_layoutnet} as $\alpha = \beta = 1$. 

\subsubsection{DuLa-Net training}

DuLa-Net uses the ADAM optimizer with $\beta_1=0.9$ and $\beta_2=0.999$. The learning rate is $0.0001$ and batch size is $8$. The training loss converges after about $120$ epochs. For feature fusion, the $\alpha$ and $\beta$ in~\eqnref{equ:fusion} is set to be $0.6$ and $3$. The $\gamma$ in~\eqnref{equ:loss_dulanet} is set to be $0.5$.

\subsection{Summarization of Modifications}

As introduced in~\secref{subsec:method} and~\secref{subsec:implementation}, we unify some of the designs and training details and propose the modified LayoutNet and DuLa-Net methods. For clarity, we summarize our modifications to LayoutNet~(denoted as \textbf{``{\layoutnetEX}"}) and DuLa-Net~(denoted as \textbf{``{\dulanetEX}"}) as follows.

\subsubsection{\layoutnetEX}

For \layoutnetEX, we use pre-trained ResNet encoder instead of SegNet encoder trained from scratch. We add random stretching data augmentation. We perform 3D layout fitting using gradient ascent optimization instead of sampling based searching scheme. We extend the equirectangular view optimization for general Manhattan layout. 

\subsubsection{\dulanetEX}

For \dulanetEX, we choose to use deeper ResNet encoders instead of the ResNet-18 one and add random stretching data augmentation.

%% file: tbl_taxonomy_1.tex
\begin{table*}[!t]
\caption{Taxonomy of key design choices of LayoutNet, DuLa-Net and HorizonNet originally proposed in \cite{zou2018layoutnet}, \cite{yang2019dula} and~\cite{sun2019horizonnet}.}\label{tab:taxonomy_1}
\begin{center}
\resizebox{1.0\textwidth}{!}{
\begin{tabular}{|c|c|c|c|c|c|c|c|c|c|c|c|c|c|}
\hline
\multirow{3}{*}{Method} & \multicolumn{3}{c|}{\multirow{2}{*}{Input}} & \multirow{3}{*}{\begin{tabular}{c} Pre- \\ process  \end{tabular}} & \multicolumn{4}{c}{Network Architecture} & \multicolumn{3}{|c}{\multirow{2}{*}{Output}} & \multicolumn{2}{|c|}{\multirow{2}{*}{Post-processing}}\\
\cline{6-9}
& \multicolumn{1}{c}{}&\multicolumn{1}{c}{}&&&\multicolumn{2}{c|}{Encoder} & \multicolumn{2}{c|}{Decoder}&\multicolumn{1}{c}{}&\multicolumn{1}{c}{}&&\multicolumn{1}{c}{}&\\
\cline{2-4}
\cline{6-14}
&\begin{tabular}{c} RGB in Equi-\\rectangular View  \end{tabular}& \begin{tabular}{c} RGB in \\Ceiling View  \end{tabular} & \begin{tabular}{c} Manhattan\\Lines Map  \end{tabular} & & SegNet & ResNet & \begin{tabular}{c} Equirectangular\\View  \end{tabular} & \begin{tabular}{c} Ceiling \\View\end{tabular} &\begin{tabular}{c} Corner\\Position  \end{tabular} & \begin{tabular}{c} Boudary\\Position  \end{tabular} & \begin{tabular}{c} Floor\\Map  \end{tabular}& \begin{tabular}{c} Equirectangular\\View  \end{tabular}& \begin{tabular}{c} Ceiling \\View\end{tabular}\\ 
\hline
\color{Purple}\textbf{LayoutNet}& \color{Purple}\CIRCLE  & &\color{Purple}\CIRCLE & \color{Purple}\CIRCLE & \color{Purple}\CIRCLE & & \color{Purple}\CIRCLE & & \color{Purple}\CIRCLE & \color{Purple}\CIRCLE & & \color{Purple}\CIRCLE &\\
\hline
\color{BrickRed}\textbf{DuLa-Net}& \color{BrickRed}\CIRCLE & \color{BrickRed}\CIRCLE & & \color{BrickRed}\CIRCLE && \color{BrickRed}\CIRCLE & \color{BrickRed}\CIRCLE & \color{BrickRed}\CIRCLE &&& \color{BrickRed}\CIRCLE && \color{BrickRed}\CIRCLE\\
\hline
\color{YellowOrange}\textbf{HorizonNet} & \color{YellowOrange}\CIRCLE&&&\color{YellowOrange}\CIRCLE&&\color{YellowOrange}\CIRCLE&\color{YellowOrange}\CIRCLE&&\color{YellowOrange}\CIRCLE&\color{YellowOrange}\CIRCLE&&&\color{YellowOrange}\CIRCLE\\
\hline
\end{tabular}
}
\end{center}
\end{table*}

%% file: fig_layoutnet_overview.tex
\begin{figure*}[!t]
\begin{center}
\includegraphics[width=0.99\linewidth]{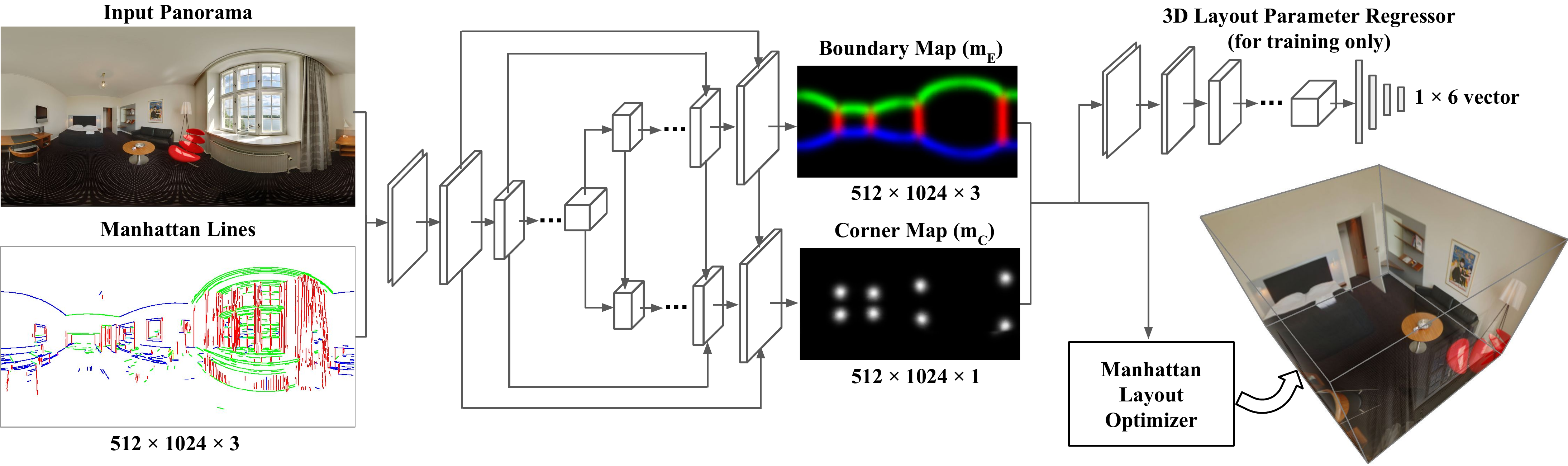}
\end{center}
\vspace{-3mm}
   \caption{Network architecture of LayoutNet. LayoutNet follows the encoder-decoder strategy. The network input is a concatenation of a single RGB panorama and Manhattan line map. The network 
   jointly predicts layout boundaries and corner positions. The 3D layout parameter loss 
   encourages predictions that maximize accuracy. The final prediction is a Manhattan layout reconstruction. Best viewed in color.}
   \vspace{-1.0em}
\label{fig:layoutnet-overview}
\end{figure*}

%% file: fig_dulanet_overview.tex
\begin{figure*}[!t]
  \centering
  \includegraphics[width=0.99\linewidth]{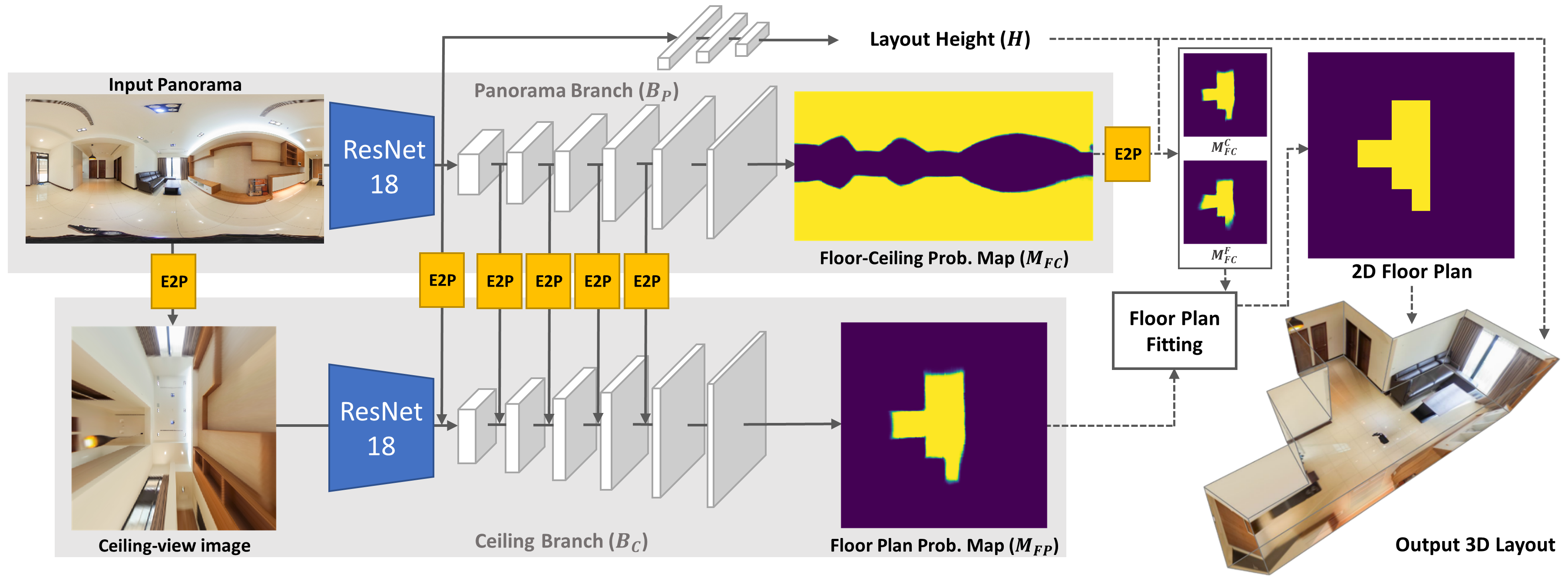}
  \caption{
  Network architecture of DuLa-Net. DuLa-Net follows the encoder-decoder scheme and consists of two branches. Given a panorama in equirectangular projection, DuLa-Net additionally creates a perspective {\em ceiling-view} image through a equirectangular-to-perspective ({\EtoP}) conversion. The panorama and the ceiling-view images are then fed to the {\em panorama-view} (upper) and {\em ceiling-view} (lower) branches. A {\EtoP}-based feature fusion scheme is employed to connect the two branches, which are jointly trained by the network to predict: 1) probability map of the floor and ceiling in panorama view, 2) a floor plan in ceiling view, and 3) a layout height. Then, the system fits a 2D Manhattan floor plan from the weighted average of the three probability maps, which is further extruded using the predicted layout height to obtain the final 3D room layout.
  }
  \label{fig:dulanet-overview}
\end{figure*}

\begin{figure*}[!t]
  \centering
  \includegraphics[width=0.99\linewidth]{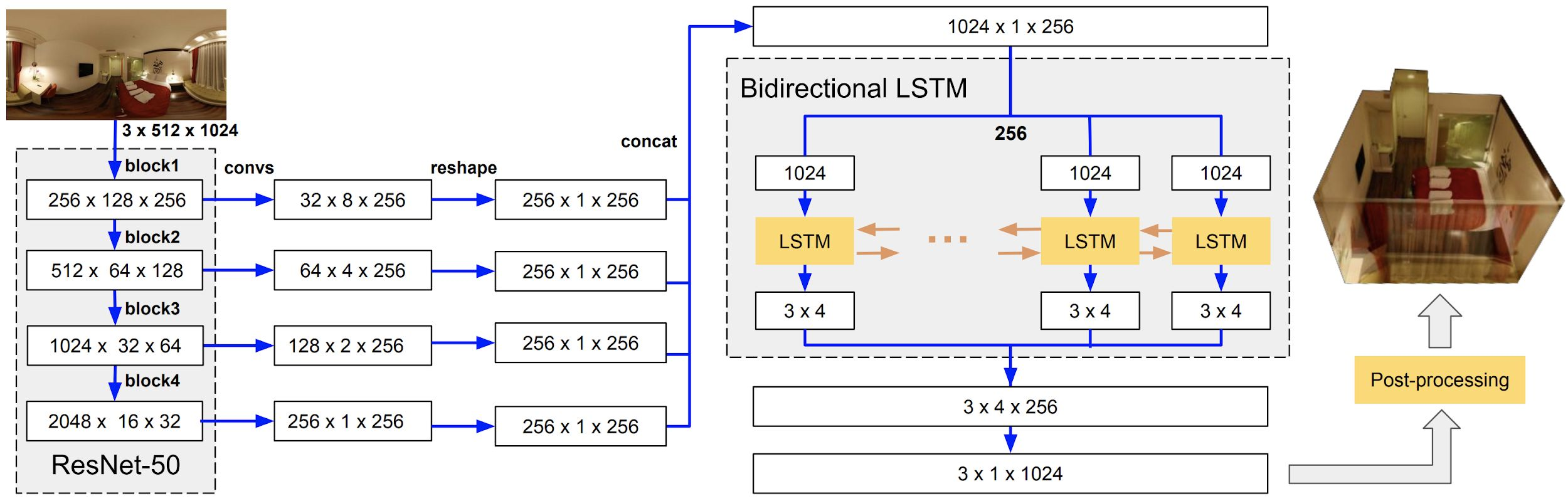}
  \caption{
  Network architecture of HorizonNet. HozionNet follows the encoder-decoder scheme. The network performs a separate convolution for each of the feature maps produced by each block of the ResNet encoder. The features are than concatenated to form the bottleneck feature. At the decoding stage, HorizonNet predicts three 1-D vectors with 1024 dimensions, representing the ceiling-wall and the floor-wall boundary position, and the existence of wall-wall boundary (or corner) of each 512x1024 image column. An RNN block is applied in the decoder to refine the vector predictions. The final post-processing step is done under ceiling view to produce the complete 3D room layout.}
  \label{fig:horizonnet-overview}
\end{figure*}

%% file: fig_floorplan_fitting.tex
\begin{figure}[!t]
    \centering
    \includegraphics[width=\columnwidth]{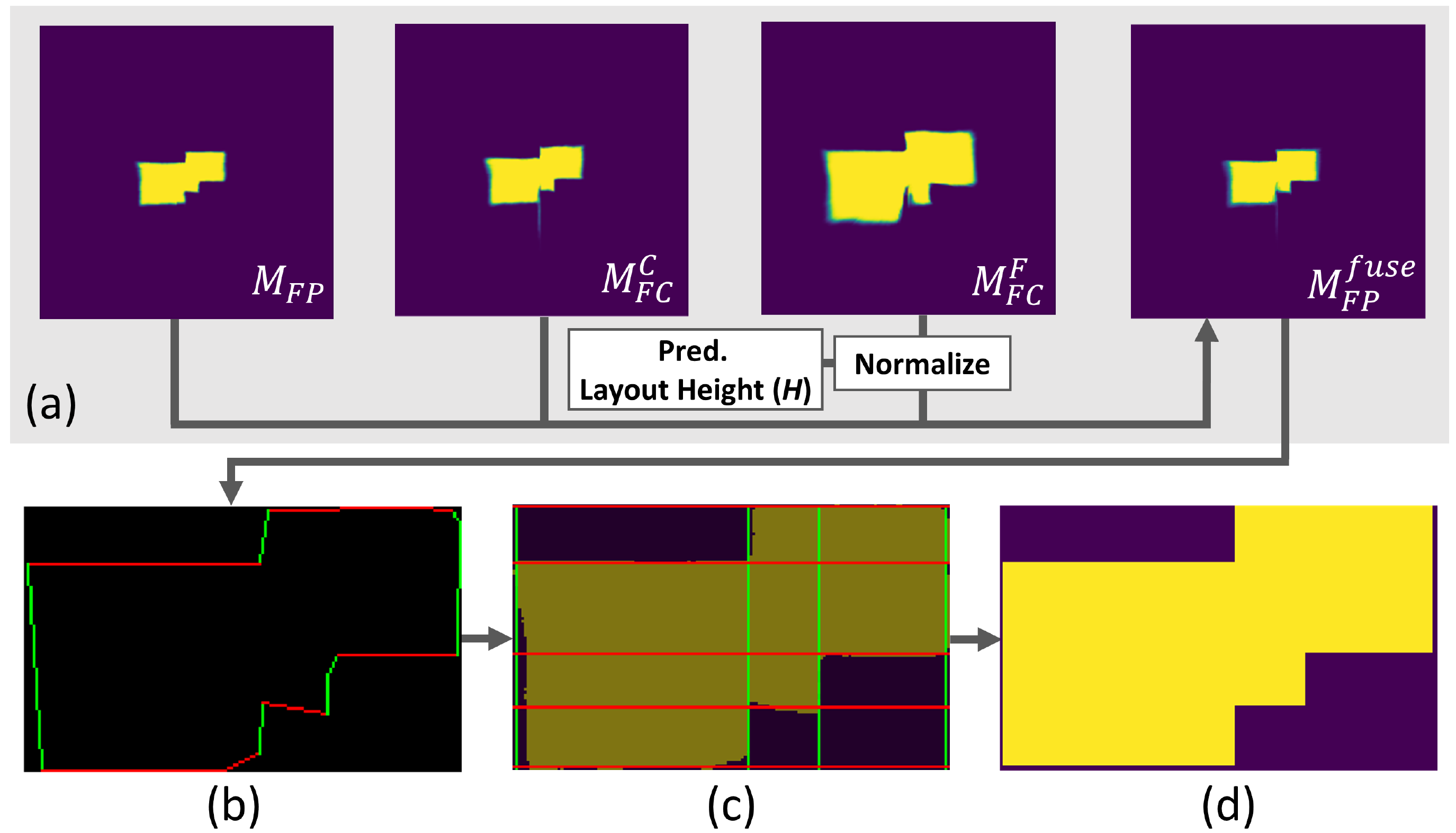}
    \caption{
    \textbf{2D floor plan fitting.} (a) The probability maps that DuLa-Net outputs are fused to a {\fpName} $M_{FP}^{fuse}$. 
    (b) Image thresholding is applied to $M_{FP}^{fuse}$ and a polygon shape is fitted to the floor plan region.
    (c) The polygon edges are regressed and clustered into two sets of horizontal lines (red) and vertical lines (green).
    (d) The final floor plane shape is defined by grids in (c) where the ratio of floor plan area is greater than 0.5.}
    \label{fig:floorplan_fitting}
\end{figure}

%% file: tbl_taxonomy_2.tex
\begin{table}[!t]
\begin{center}
\caption{Taxonomy of network training details~(data augmentation details) of LayoutNet, DuLa-Net and HorizonNet originally proposed in ~\cite{zou2018layoutnet}, \cite{yang2019dula} and~\cite{sun2019horizonnet}. Note that DuLa-Net uses horizontal rotatations for 0$^\circ$, 90$^\circ$, 180$^\circ$, and 270$^\circ$ only instead of all 0-360$^\circ$ to generate axis-aligned floor maps, so we mark it as a half circle for difference. }\label{tab:taxonomy_2}
\resizebox{0.49\textwidth}{!}{
\begin{tabular}{|c|c|c|c|c|c|}
\hline
Method & \begin{tabular}{c} Left-right\\Flipping  \end{tabular}  &  \begin{tabular}{c} Horizontal\\Rotation  \end{tabular}   & \begin{tabular}{c} Luminance\\Change  \end{tabular}& \begin{tabular}{c} Ground Truth\\Smoothing  \end{tabular}& \begin{tabular}{c} Random\\Stretching  \end{tabular}\\
\hline
\color{Purple}\textbf{LayoutNet}&\color{Purple}\CIRCLE&\color{Purple}\CIRCLE&\color{Purple}\CIRCLE&\color{Purple}\CIRCLE&\\
\hline
\color{BrickRed}\textbf{DuLa-Net}&\color{BrickRed}\CIRCLE &\color{BrickRed}\LEFTcircle& \color{BrickRed}\CIRCLE& &\\
\hline
\color{YellowOrange}\textbf{HorizonNet}&\color{YellowOrange}\CIRCLE&\color{YellowOrange}\CIRCLE&\color{YellowOrange}\CIRCLE&\color{YellowOrange}\CIRCLE&\color{YellowOrange}\CIRCLE\\
\hline
\end{tabular}
}
\end{center}
\end{table}

%% file: tbl_cuboid_pano.tex
\begin{table*}[!t]
\caption{Quantitative comparison for cuboid layout estimation with a unified encoder on the PanoContext dataset. Bold numbers indicate the best performance. 
}
\begin{center}
\resizebox{1.0\textwidth}{!}{
\begin{tabular}{|c|c|c|c|c|c|c|c|c|c|}
\hline
\multirow{2}{*}{Encoder} & \multicolumn{3}{c}{3D IoU~(\%)} & \multicolumn{3}{|c|}{Corner Error~(\%)} & \multicolumn{3}{c|}{Pixel Error~(\%)} \\
\cline{2-10}
& {\layoutnetEX} & {\dulanetEX} & HorizonNet & {\layoutnetEX} & {\dulanetEX} & HorizonNet & {\layoutnetEX} & {\dulanetEX} & HorizonNet\\
\hline
ResNet-18 & 84.13& 82.43 & 80.27 & 0.65 & 0.83 & 0.83 & 1.92 & 2.55& 2.44\\
ResNet-34 & \textbf{85.02}& 83.41 & 81.30 & \textbf{0.63} & 0.82 & 0.76& \textbf{1.79} & 2.54 & 2.22\\
ResNet-50 & 82.44& 83.77& 82.63 & 0.75 & 0.81 & 0.74 & 2.22 & 2.43&2.17\\
\hline
\end{tabular}
}
\end{center}
\label{tab:cuboid_pano}
\end{table*}

%% file: tbl_cuboid_stdn.tex
\begin{table*}[!t]
\caption{Quantitative comparison for cuboid layout estimation with a unified encoder on Stanford 2D-3D dataset. Bold numbers indicate the best performance. 
}
\begin{center}
\resizebox{1.0\textwidth}{!}{
\begin{tabular}{|c|c|c|c|c|c|c|c|c|c|}
\hline
\multirow{2}{*}{Encoder} & \multicolumn{3}{c}{3D IoU~(\%)} & \multicolumn{3}{|c|}{Corner Error~(\%)} & \multicolumn{3}{c|}{Pixel Error~(\%)} \\
\cline{2-10}
& {\layoutnetEX} & {\dulanetEX} & HorizonNet & {\layoutnetEX} & {\dulanetEX} & HorizonNet & {\layoutnetEX} & {\dulanetEX} & HorizonNet\\
\hline
ResNet-18 & 83.53 & 84.93 & 80.59 & 0.77 & 0.74 & 0.82 & 2.30 & 2.56 & 2.72\\
ResNet-34 & 84.17 & 86.45 & 80.44 & 0.71 & \textbf{0.66} & 0.78 & \textbf{2.04} & 2.43 & 2.65\\
ResNet-50 & 82.66 & \textbf{86.60} & 82.72& 0.83 & 0.67 & 0.69 & 2.59 &2.48 &2.27\\
\hline
\end{tabular}
}
\end{center}
\label{tab:cuboid_stdn}
\end{table*}

%% file: Performance_comparison.tex
\section{Experiments and Discussions}
In this section, we evaluate the performance of {\layoutnetEX}, {\dulanetEX} and HorizonNet introduced in~\secref{sec:network}. We describe the evaluation metrics in~\secref{exp:metric} and compare the methods on PanoContext dataset and Stanford 2D-3D dataset for cuboid layout reconstruction in~\secref{exp:cuboid}. We evaluate performance on {\datasetName} for general Manhattan layout estimation in~\secref{exp:matterport}. Finally, based on the experiment results, we discuss the advantages and disadvantages of each method in~\secref{exp:discuss}. 

\input{fig_visual_result_pano.tex}

\subsection{Evaluation Setup}
\label{exp:metric}
We use the following five standard evaluation metrics: 
\begin{itemize}
    \item \textbf{Corner error}, which is the $L2$ distance between the predicted layout corners and the ground truth under equirectangular view. The error is normalized by the image diagonal length and averaged across all images.
    \item \textbf{Pixel error}, which is the pixel-wise semantic layout prediction (wall, ceiling, and floor) accuracy compared to the ground truth. The error is averaged across all images.
    \item \textbf{3D IoU}, defined as the volumetric intersection over union between the predicted 3D layout and the ground truth. 
    The result is averaged over all the images.
    \item \textbf{2D IoU}, defined as the pixel-wise intersection over union between predicted layout under ceiling view and the ground truth. 
    The result is averaged over all the images.
    \item \textbf{rmse}, defined as the root mean squared error between predicted layout depth $\hat{d}$ and the ground truth $d$: \\
    $\sqrt{\frac1{|d|} \sum_{p\in d}{(d_p - \hat{d}_p)^2}}$ where p represents every pixel in the depth. We use the true camera height, which is 1.6 for each image, to generate the predicted depth map.
    The result is averaged over all the images.
    \item $\boldsymbol{\delta_i}$, defined as the percentage of pixels where the ratio~(or its reciprocal) between the prediction and the label is within a threshold of 1.25:
    $\frac1{|d|} \sum_{p\in d} \mathbf{1}[\max{(\frac{d_p}{\hat{d_p}},\frac{\hat{d_p}}{d_p})} < 1.25]$. 
\end{itemize}
We use corner error, pixel error, and 3D IoU to evaluate performance of cuboid layout reconstruction.
For general Manhattan layout reconstruction, since the predicted layout shape can be different from the ground truth shape, we use 3D IoU, 2D IoU and depth measurements (\ie rmse and $\delta_1$) for evaluation.

\label{sec:experiment}

\subsection{Performance on PanoContext and Stanford 2D-3D}
\label{exp:cuboid}
In this experiment, we evaluate the performance of {\layoutnetEX}, {\dulanetEX}, and HorizonNet on the PanoContext dataset and Stanford 2D-3D dataset, which is comprised of cuboid layouts. For all three methods, we used a unified (ResNet) encoder and analyzed the performance of using different post-processing steps.
%
%

\paragraph{Dataset setting.} For the evaluation on PanoContext dataset, we use both the training split of PanoContext dataset and the whole Stanford 2D-3D dataset for training and vice versa for the evaluation on Stanford 2D-3D dataset. The split for validation and testing of each dataset is reported in~\secref{sec:dataset}. We use the same dataset setting for all three methods.

\paragraph{Qualitative results.} We show in~\figref{fig:pano} the qualitative results of the experiments on PanoContext dataset and Stanford 2D-3D dataset. 
All methods offer similar accuracy. LayoutNet v2 slightly outperforms on PanoContext and offers more robustness to occlusion from foreground objects, while DuLa-Net v2 outperforms in two of three metrics for Stanford 2D-3D as shown in Table~\ref{tab:cuboid_pano} and Table~\ref{tab:cuboid_stdn}.

\subsubsection{Evaluation on Unified Encoder}
\label{text:exp_encoder}
\tabref{tab:cuboid_pano} and~\tabref{tab:cuboid_stdn} show the performance for {\layoutnetEX}, {\dulanetEX} and HorizonNet on PanoContext dataset and Stanford 2D-3D dataset, respectively. In each row, we report performance by using ResNet-18, ResNet-34, and ResNet-50 encoders respectively. For both {\dulanetEX} and HorizonNet, using ResNet-50 obtains the best performance, indicating that deeper encoder can better capture layout features. For {\layoutnetEX}, we spot a performance drop with ResNet-50, this is mainly due to the smaller number of batch size (we use 2 in experiment, which is the maximum available number to run on a single GPU of 12GB) that leads to unstable training of the batch normalization layer in ResNet encoder. We expect an better performance of {\layoutnetEX} with ResNet-50 by training on a GPU with a larger memory, but we consider it as an unfair comparison with the other two methods since the hardware setup is different. In general, {\layoutnetEX} with ResNet-34 outperforms all other methods on PanoContext dataset and obtains lowest pixel error on Stanford 2D-3D dataset. {\dulanetEX}, on the other hand, shows the best 3D IoU and corner error on Stanford 2D-3D dataset.
Note that the reported number for HorizonNet with ResNet-50 is slightly lower than that reported in the original paper. This is attributed to the difference in the training dataset, \ie the authors used both the training split of PanoContext dataset and Stanford 2D-3D dataset for training. We thus retrain the HorizonNet using our training dataset setting for a fair comparison.

\subsubsection{Ablation Study}
\label{text:abla}
We show in~\tabref{tab:cuboid_abla} the ablation study of different design choices of {\layoutnetEX} on the best performing PanoContext dataset. The first row shows the performance reported in~\cite{zou2018layoutnet}. The proposed {\layoutnetEX} with ResNet encoder, modified data augmentation and post-processing step boosts the overall performance by a large margin ($\sim 10\%$ in 3D IoU).
A large performance drop is observed when training the model from scratch (w/o ImageNet pre-training).
%
%
%
Using gradient ascent for post-processing contributes the most to the performance boost (w/o gradient ascent), while adding random stretching data augmentation contributes less (w/o random stretching).
Freezing batch normalization layout when training end-to-end can avoid unstable training of this layer when the batch size is small (w/o freeze bn layer).
Including all modifications together achieves the best performance. 

We show in~\tabref{tab:cuboid_abla_dula} the ablation study for {\dulanetEX} on the Stanford 2D-3D dataset.
We obtain a performance boost of $5\%$ in 3D IoU when comparing with the original model~\cite{yang2019dula} by using a deeper ResNet encoder (ResNet-50 vs. ResNet-18). Similar to {\layoutnetEX}, using the random stretching data augmentation (w/o random stretching) improves the performance only marginally.

\input{tbl_cuboid_abla.tex}
\input{tbl_cuboid_abla_dula.tex}
\input{tbl_time.tex}

\paragraph{Comparison with different post-processing steps.}
In this experiment, we compare the performance of {\layoutnetEX} while using the post-processing steps of {\dulanetEX} and HorizonNet, and combining its own optimization step with additional semantic loss, respectively.
The post-processing step of HorizonNet utilizes predicted layout boundaries and corner positions in each image column, which can be easily converted from the output of {\layoutnetEX}. To adapt {\dulanetEX}'s post-processing step, we train {\layoutnetEX} to predict the semantic segmentation (\ie wall probability map) under equirectangular view as an additional channel in the boundary prediction branch. Then, we use the predicted floor-ceiling probability map as input to the post-processing step of {\dulanetEX}. Alternatively, we can also incorporate the predicted wall probability map into the layout optimization of {\layoutnetEX}. We add an additional loss term to~\eqnref{equ:layotnetopt} for the average per-pixel value enclosed in the wall region of the predicted probability map with a threshold of 0.5. We set the semantic term weights to 0.3 for grid search in the validation set.
As reported in~\tabref{tab:cuboid_abla} (row~6-8), together with {\layoutnetEX}'s neural network, a post-processing under equirectangular view performs better than the one under ceiling view. We found that the additional semantic optimization did not improve the post-processing step under equirectangular view. This is because the jointly predicted semantic segmentation is not that accurate, achieving only 2.59\% pixel error compared with the 1.79\% pixel error by our proposed LayoutNet v2.

Another interesting study is to see whether the performance of {\dulanetEX} and HorizonNet will be affected by using the post-processing step that works on the equirectangular view. However, it is not clear how to convert from their output probability maps to layout boundaries and corner positions, which are the required input for {\layoutnetEX}'s post-processing step.

\input{fig_visual_result_matterport.tex}

\subsubsection{Timing Statistics}
We show in~\tabref{tab:time} the timing performance of {\layoutnetEX} with ResNet-34 encoder, {\dulanetEX} with ResNet-50 encoder, and HorizonNet with ResNet-50 encoder. We report the computation time of HorizonNet with RNN refinement branch. Note that HorizonNet without RNN only costs 8ms for network prediction but produces less accurate result compared with other approaches. We report average time consumption for a single forward pass of the network and the post-processing step.


\input{fig_eval_confuse_mat.tex}

\subsection{Performance on {\datasetName}}
\label{exp:matterport}
In this experiment, we compare the performance of three methods on estimating the general Manhattan layouts using the {\datasetName} dataset.
For a detailed evaluation, we report the performance for layouts of different complexity. We  categorize each layout shape according to the number of floor plan corners in the ceiling view, e.g. a cuboid has 4 corners, an ``L''-shape has 6 corners, and a ``T''-shape has 8 corner.
The dataset split used for training/validation/testing is reported in~\secref{sec:dataset}.


\paragraph{Qualitative results.}
\label{exp:matterport_quali}
\figref{fig:matterport_eval} shows the qualitative comparisons of the three methods.
All three methods have similar performance when the room shape is simpler, such as cuboid and `L''-shape rooms. For more complex room shapes, HorizonNet is capable of estimating thin structures like the walls as shown in~\figref{fig:matterport_eval} (6th row, 1st column), but could also be confused by the reflected room boundaries in the mirror as shown in~\figref{fig:matterport_eval} (6th row, 4th column).
{\layoutnetEX} tends to ignore the thin layout structures like the bumped out wall as shown in~\figref{fig:matterport_eval} (7th row, 1st column).
{\dulanetEX} is able to estimate the occluded portion of the scene, utilizing cues from the 2D ceiling view as shown in~\figref{fig:matterport_eval} (8th row, 2nd column), but could also be confused by ceiling edges as shown in~\figref{fig:matterport_eval} (8th row, last column).


\input{tbl_eval_matterport.tex}


\paragraph{Quantitative Evaluation.} \tabref{tab:eval_matterport} shows the quantitative comparison of three methods on estimating general Manhattan layout using the {\datasetName} dataset.
We consider the 3D IoU, 2D IoU and two depth accuracy measurements (\ie rmse and $\delta_1$) for the performance evaluation.
Overall, among the three methods, HorizonNet shows the best performance while {\layoutnetEX} has similar performance on 2D IoU and 3D IoU with cuboid room shape. {\dulanetEX} performs better than {\layoutnetEX} for non-cuboid shapes, while being slightly worse than HorizonNet. Although these three methods show competitive performance in the overall 2D and 3D IoU metric, the performance gap in the depth metrics is more obvious. This is because the depth metrics can quantify the detailed local geometric differences: predicting an ``L''- shape room with a small concave corner as a cuboid room can have less impact on 2D or 3D IoU, but the depth error will increase. This indicates the value of our newly proposed metrics.

\subsection{Discussions}
\label{exp:discuss}
\subsubsection{Why do {\layoutnetEX} and HorizonNet perform differently on different datasets?}\label{exp:discuss_conf}
On PanoContext dataset and Stanford 2D-3D dataset, {\layoutnetEX} outperforms the other two methods. However, on {\datasetName} dataset, HorizonNet stands to be the clear winner. We believe this is due to the different design of network decoder and the different representation of network's outputs, making each method performs differently for cuboid layout and non-cuboid layout, as discussed below.

%

{\layoutnetEX} relies more on the global room shape context, \ie it can predict one side of the wall given the prediction of the other three walls. This is benefited from the two-branch network prediction of room boundaries and corners, and the corner prediction is guided by the room boundaries: boundaries will also get gradients from error predicted corners during training. However, because of the reliance on the global context for {\layoutnetEX}, it is harder to generalize the reasoning process from one layout type to another. For example, learning how to use the global context on images with cuboid layout might not help in a room with 16 corners. 
This gap is lesser for HorizonNet and {\dulanetEX} since they predict local outputs that can be generalized to arbitrary layouts. This is because HorizonNet and {\dulanetEX} emphasize more on local edge and corner responses, \eg predict whether this column has a corner, and the position of floor and ceiling in this column. A direct evidence is that, by training on Stanford 2D-3D dataset which has all cuboid shapes, {\layoutnetEX} predicts cuboid shape only, while HorizonNet has $10\%$ non-cuboid outputs. These characteristics are also reflected in the qualitative results shown in~\figref{fig:matterport_eval}. As we discussed in~\secref{exp:matterport_quali}, {\layoutnetEX} often misses thin layout structures such as pipes, while HorizonNet can be more sensitive to those thin structures. 
We also show in~\figref{fig:cf} the confusion matrix on correctly estimating the number of corners of the 3D layouts for each method.
%
%
For the cuboid layout~(4 corners), {\layoutnetEX} shows the highest recall rate. However, {\layoutnetEX} also tends to predict some non-cuboid layouts~(\eg 6 corners, 8 corners, 10 corners) to be cuboid. 
On the other hand, {\dulanetEX} and HorizonNet shows better and comparable performance for estimating the non-cuboid room layouts. Therefore, the error in layout type prediction is the major cause of error for {\layoutnetEX} in 3D reconstruction on the MatterportLayout dataset. 


Moreover, HorizonNet differs from {\layoutnetEX} and {\dulanetEX} in the decoder architecture. HorizonNet uses a 1D RNN, while {\layoutnetEX} and {\dulanetEX} use 2D convolutions. We believe that this is also one of the reasons why HorizonNet's performance in the PanoContext dataset lags behind {\layoutnetEX} and {\dulanetEX}: 2D convolutions can better localize low level details like corners, lines and curves.

\subsubsection{Analysis and Future Improvements for {\dulanetEX}}
{\dulanetEX} is sensitive to the parameter of FOV (introduced in the E2P projection in~\secref{subsec:preprocess}). A smaller FOV (\eg $160^{\circ}$) can lead to higher quality predictions for most of the rooms, but some larger rooms could be clipped by the image plane after projection. A larger FOV (\eg $165^{\circ}$, $171^{\circ}$) could produce fewer clipped rooms after projection, but the prediction quality for some rooms may decrease, due to the down-scaled ground truth 2D floor plan in ceiling view. In this paper, we use the setting of FOV=$160^{\circ}$, but we suggest to improve the prediction quality by combining the prediction of multiple networks trained with different FOVs in the future work. To give an idea of the potential improvement, we report the numbers for {\datasetName} dataset by removing the rooms that are too big to be clipped by the boundary of the projection under the setting of FOV=$160^{\circ}$. For this case, the 3D IoU improves from 74.53 to 76.82.





%% file: fig_visual_result_pano.tex
\begin{figure*}[!t]
\includegraphics[width=\textwidth]{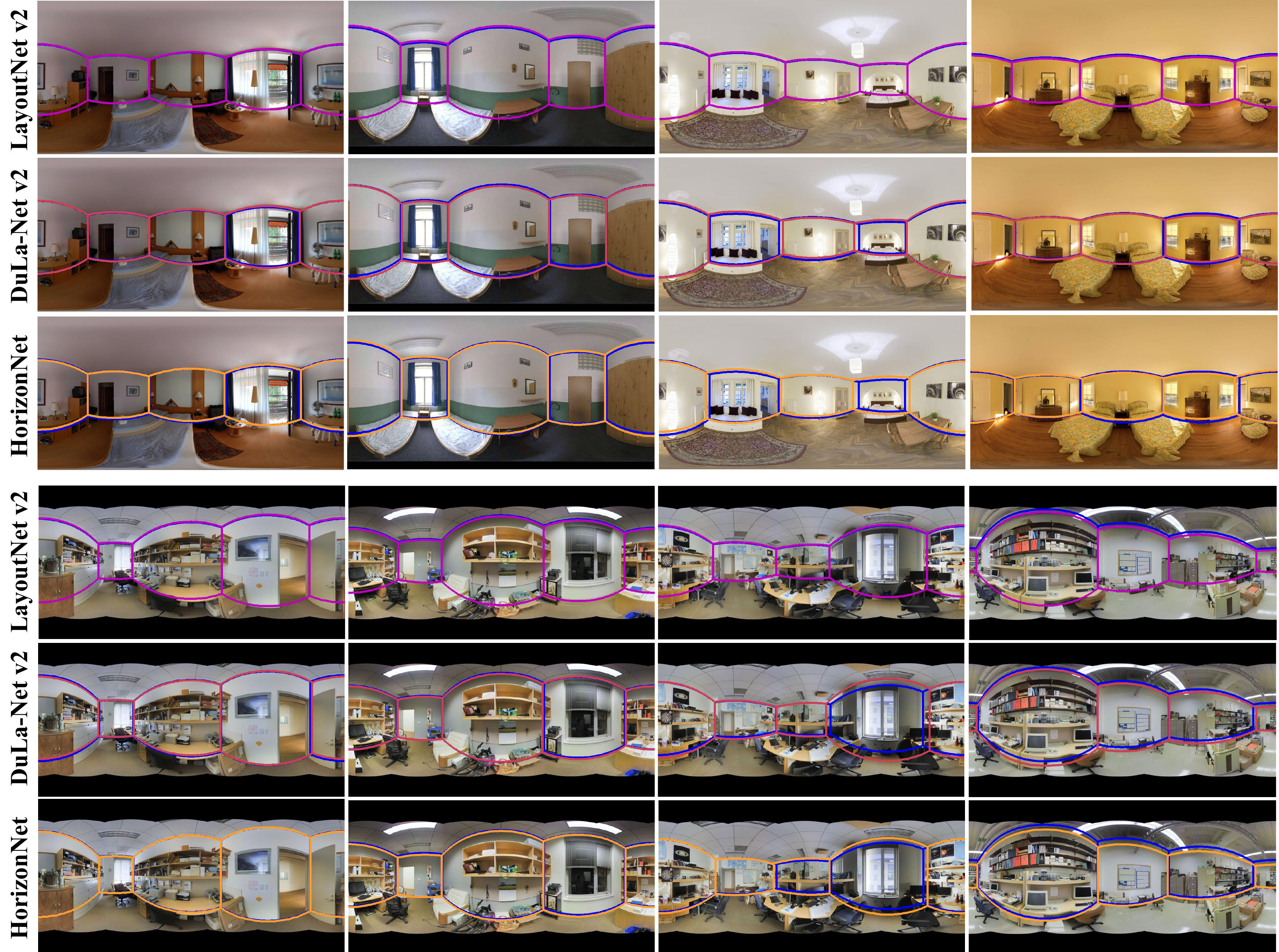}
\caption{Qualitative comparison on cuboid layout estimation. We show results selected randomly from the experiment of testing {\layoutnetEX}, {\dulanetEX} and HorizonNet on PanoContext dataset (top) and Stanford 2D-3D dataset (bottom). In each example, we show the predicted layout~(LayoutNet v2: purple, DuLa-Net v2: red, HorizonNet: orange) and the ground truth~(blue) under equirectangular view.}
\label{fig:pano}       
\end{figure*}

%% file: tbl_cuboid_abla.tex
\begin{table}
\begin{center}
\caption{Ablation study of {\layoutnetEX} for cuboid layout estimation on the PanoContext dataset. 
We report the best performing {\layoutnetEX} with ResNet-34 encoder (bold numbers). We show in the first row the performance of original LayoutNet~\cite{zou2018layoutnet}, and row 2-5 the ablation study for {\layoutnetEX}. Row 7-9 show the performance comparison using different post-processing steps.
}\label{tab:cuboid_abla}
\resizebox{0.49\textwidth}{!}{
\begin{tabular}{|c|c|c|c|c|}
\hline
Method & 3D IoU~(\%) & \begin{tabular}{c} Corner \\ error~(\%)  \end{tabular}  & \begin{tabular}{c} Pixel \\ error~(\%)  \end{tabular}\\
\hline\hline
LayoutNet~\cite{zou2018layoutnet}& 75.12& 1.02 & 3.18\\
\hline
w/o ImageNet pre-train & 78.71 & 0.89 & 2.57\\
w/o gradient ascent & 83.60&0.73&2.12\\
w/o freeze bn layer& 83.98 & 0.70 & 2.01\\
w/o random stretching& 83.97 & 0.65 & 1.92\\
\hline
LayoutNet v2 & \textbf{85.02}& \textbf{0.63} & \textbf{1.79}\\
\hline
\hline
w/ DuLa-Net post-proc & 81.45&0.90&2.73\\
w/ HorizonNet post-proc &82.70 &0.77&2.15\\
w/ Semantic post-proc & 84.35&0.65&1.96\\
\hline
\end{tabular}
}
\end{center}
\end{table}

%% file: tbl_cuboid_abla_dula.tex
\begin{table}
\begin{center}
\caption{Ablation study of {\dulanetEX} for cuboid layout estimation on the Stanford 2D-3D dataset. 
We report the best performing {\dulanetEX} with ResNet-50 encoder (bold numbers).}
\label{tab:cuboid_abla_dula}
\resizebox{0.49\textwidth}{!}{
\begin{tabular}{|c|c|c|c|c|}
\hline
Method & 3D IoU~(\%) & \begin{tabular}{c} Corner \\ error~(\%)  \end{tabular}  & \begin{tabular}{c} Pixel \\ error~(\%)  \end{tabular}\\
\hline\hline
DuLa-Net~\cite{yang2019dula} &81.59&1.06&3.06\\
\hline
w/o random stretching& 85.03 & 0.94 & 2.85\\
DuLa-Net v2 & \textbf{86.60} & \textbf{0.67} & \textbf{2.48}\\
\hline
\end{tabular}
}
\end{center}
\end{table}

%% file: tbl_time.tex
\begin{table}[!b]
\caption{Timing comparison. We report average time consumption for a single forward pass of the neural network and the post-processing step respectively. We report performance of {\layoutnetEX} with ResNet-34 encoder, {\dulanetEX} with ResNet-50 encoder and HorizonNet with ResNet-50 encoder.}\label{tab:time}
\begin{center}
\resizebox{0.49\textwidth}{!}{
\begin{tabular}{|c|c|c|}
\hline
Method & Optimization avg CPU Time (ms) & Network avg. GPU time (ms)\\
\hline
{\layoutnetEX} & 1222 & \textbf{30}\\
{\dulanetEX} & 35& 42\\
HorizonNet &\textbf{12} & 50\\
\hline
\end{tabular}
}
\end{center}
\end{table}

%% file: fig_visual_result_matterport.tex
\begin{figure*}[!t]
 \includegraphics[width=\textwidth]{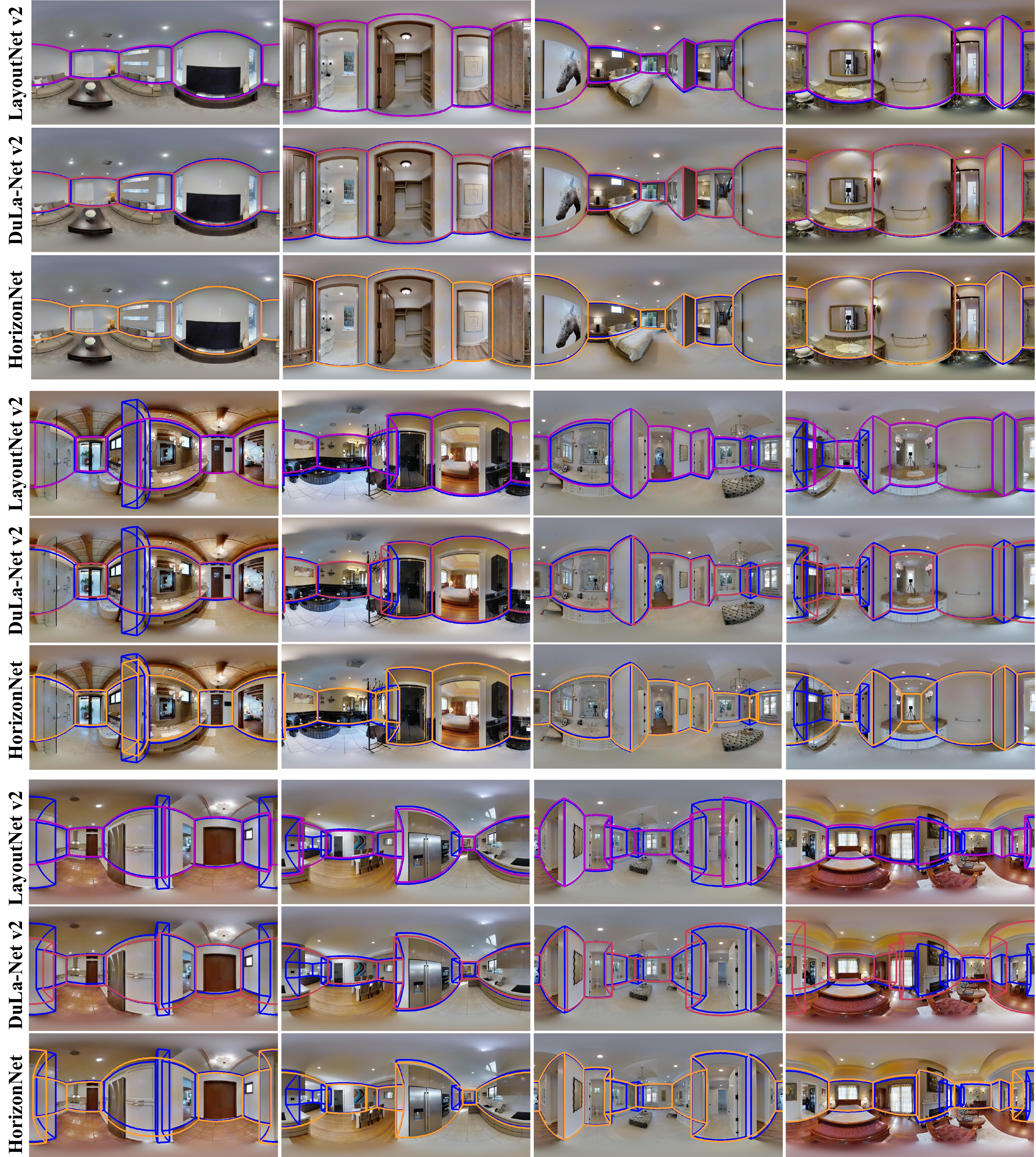}
\caption{Qualitative comparison on the {\datasetName} dataset. We show the 3D room layouts with various complexities estimated by {\layoutnetEX}, {\dulanetEX} and HorizonNet. In each example, we show the predicted layout~(LayoutNet v2: purple, DuLa-Net v2: red, HorizonNet: orange) and the ground truth~(blue) under equirectangular view. 
}
\label{fig:matterport_eval}       
\end{figure*}

%% file: fig_eval_confuse_mat.tex
\begin{figure*}[!t]
  \centering
  \includegraphics[width=\textwidth]{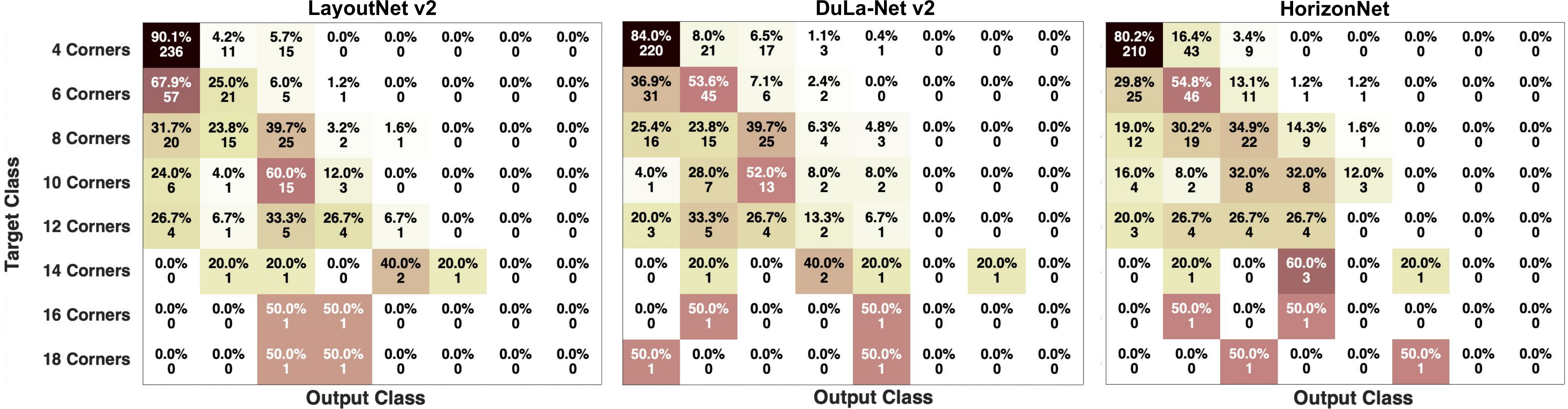}
  \caption{The confusion matrix for the performance evaluation of {\layoutnetEX}, {\dulanetEX}, and HorizonNet on correctly estimating the number of layout corners on the MatterportLayout dataset. We calculate the number of corners of each predicted 3D layout~(after post-processing step) by projecting the layout onto the ceiling view and record the number of wall-wall intersections in 2D. This figure illustrates the discussions of different performance of the three methods on different datasets~(\secref{exp:discuss_conf}). {\layoutnetEX} shows the highest recall rate for the cuboid layout~(4 corners), but tends to predict some non-cuboid layouts~(\eg 6 corners, 8 corners, 10 corners) to be cuboid. 
  On the other hand, {\dulanetEX} and HorizonNet show better and comparable performance for estimating the non-cuboid room layouts. These observations support our conjecture that the error in layout type prediction~(due to the reliance on global context as discussed in~\secref{exp:discuss_conf}) is the major cause of error for {\layoutnetEX} for 3D reconstruction on the MatterportLayout dataset.
  }
  \label{fig:cf}
\end{figure*}

%% file: tbl_eval_matterport.tex
\begin{table}
\caption{Quantitative evaluation of {\layoutnetEX}, {\dulanetEX}, and HorizonNet on the general Manhattan layout estimation using the {\datasetName} dataset. From top to bottom shows results evaluated by 3D IoU, 2D IoU, and depth measurements, RMSE and $\delta_1$, respectively. We report the overall performance as well as those on each layout type respectively. Bold numbers indicate the best performance.}
\label{tab:eval_matterport}
\begin{center}
\resizebox{0.49\textwidth}{!}{
\begin{tabular}{|c|c|g|c|c|c|}
\hline
\multicolumn{6}{|c|}{Metric: 3D IoU ($\%$)} \\
\hline
Method & Overall & 4 corners & 6 corners & 8 corners & $\geqslant$ 10 corners\\
\hline\hline
{\layoutnetEX} &75.82 & 81.35 & 72.33 & 67.45 & 63.00\\
{\dulanetEX} & 75.05 & 77.02 & 78.79 & 71.03 & 63.27\\
HorizonNet & \textbf{79.11} & \textbf{81.88} & \textbf{82.26} & \textbf{71.78} & \textbf{68.32}\\
\hline \hline
\multicolumn{6}{|c|}{Metric: 2D IoU ($\%$)} \\
\hline
Method & overall & 4 corners & 6 corners & 8 corners & $\geqslant$ 10 corners\\
\hline\hline
{\layoutnetEX} & 78.73 & 84.61 & 75.02 & 69.79 & 65.14\\
{\dulanetEX} & 78.82 & 81.12 & 82.69 & \textbf{74.00} & 66.12\\
HorizonNet & \textbf{81.71} & \textbf{84.67} & \textbf{84.82} & 73.91 & \textbf{70.58}\\
\hline \hline
\multicolumn{6}{|c|}{Metric: RMSE} \\
\hline
Method & overall & 4 corners & 6 corners & 8 corners & $\geqslant$ 10 corners\\
\hline\hline
{\layoutnetEX} & 0.258 & 0.212& 0.287 & 0.303 & 0.396\\
{\dulanetEX} & 0.291& 0.255 & 0.237 & 0.281 & 0.589\\
HorizonNet & \textbf{0.197}& \textbf{0.166} & \textbf{0.173} & \textbf{0.243} & \textbf{0.345}\\
\hline \hline
\multicolumn{6}{|c|}{Metric: $\delta_1$} \\
\hline
Method & overall & 4 corners & 6 corners & 8 corners & $\geqslant$ 10 corners\\
\hline\hline
{\layoutnetEX} & 0.871 & 0.897 & 0.827 & 0.877 & 0.800\\
{\dulanetEX} & 0.818 & 0.818 & 0.859 & 0.823 & 0.741\\
HorizonNet & \textbf{0.929}& \textbf{0.945} & \textbf{0.938} & \textbf{0.903} & \textbf{0.861}\\
\hline
\end{tabular}
}
\end{center}
\end{table}

\if 1
\begin{table}
\caption{3D IoU of reconstructed 3D layout on Matterport3D dataset. We report the overall performance and the performance for each type of the layout shapes respectively.}\label{tab:general_3diou}
\begin{center}
\resizebox{0.49\textwidth}{!}{
\begin{tabular}{|c|c|g|c|c|c|}
\hline
Method & Overall & 4 corners & 6 corners & 8 corners & $\geqslant$ 10 corners\\
\hline\hline
LayoutNet v2 &75.82 & 81.35 & 72.33 & 67.45 & 63.00\\
DuLa-Net v2 & 75.05 & 77.02 & 78.79 & 71.03 & 63.27\\
HorizonNet & \textbf{79.11} & \textbf{81.88} & \textbf{82.26} & \textbf{71.78} & \textbf{68.32}\\
\hline
\end{tabular}
}
\end{center}
\end{table}

\begin{table}
\caption{2D IoU of reconstructed floor plan under ceiling view on Matterport3D dataset. We report the overall performance and the performance for each type of the layout shapes respectively.}\label{tab:general_2diou}
\begin{center}
\resizebox{0.49\textwidth}{!}{
\begin{tabular}{|c|c|g|c|c|c|}
\hline
Method & overall & 4 corners & 6 corners & 8 corners & $\geqslant$ 10 corners\\
\hline\hline
LayoutNet v2 & 78.73 & 84.61 & 75.02 & 69.79 & 65.14\\
DuLa-Net v2 & 78.82 & 81.12 & 82.69 & \textbf{74.00} & 66.12\\
HorizonNet & \textbf{81.71} & \textbf{84.67} & \textbf{84.82} & 73.91 & \textbf{70.58}\\
\hline
\end{tabular}
}
\end{center}
\end{table}

\begin{table}
\caption{Layout depth estimation error~(RMSE) under equirectangular view on Matterport3D dataset. We report the overall performance and the performance for each type of the layout shapes respectively.}\label{tab:general_rmse}
\begin{center}
\resizebox{0.49\textwidth}{!}{
\begin{tabular}{|c|c|g|c|c|c|}
\hline
Method & overall & 4 corners & 6 corners & 8 corners & $\geqslant$ 10 corners\\
\hline\hline
LayoutNet v2 & 0.258 & 0.212& 0.287 & 0.303 & 0.396\\
DuLa-Net v2 & 0.291& 0.255 & 0.237 & 0.281 & 0.589\\
HorizonNet & \textbf{0.197}& \textbf{0.166} & \textbf{0.173} & \textbf{0.243} & \textbf{0.345}\\
\hline
\end{tabular}
}
\end{center}
\end{table}

\begin{table}
\caption{Layout depth estimation error~($\delta_1$) under equirectangular view on Matterport3D dataset. We report the overall performance and the performance for each type of the layout shapes respectively.}\label{tab:general_delta}
\begin{center}
\resizebox{0.49\textwidth}{!}{
\begin{tabular}{|c|c|g|c|c|c|}
\hline
Method & overall & 4 corners & 6 corners & 8 corners & $\geqslant$ 10 corners\\
\hline\hline
LayoutNet v2 & 0.871 & 0.897 & 0.827 & 0.877 & 0.800\\
DuLa-Net v2 & 0.818 & 0.818 & 0.859 & 0.823 & 0.741\\
HorizonNet & \textbf{0.929}& \textbf{0.945} & \textbf{0.938} & \textbf{0.903} & \textbf{0.861}\\
\hline
\end{tabular}
}
\end{center}
\end{table}
\fi

%% file: Conclusion.tex
\section{Conclusions and Future Work} 
\label{sec:conclusion}
In this paper, we provide a thorough analysis of the three state-of-the-art methods for 3D Manhattan layout reconstruction from a single RGB indoor panoramic image, namely, LayoutNet, DuLa-Net, and HorizonNet. We further propose the improved version called {\layoutnetEX} and {\dulanetEX}, which incorporate certain advantageous components from HorizonNet. 
LayoutNet v2 performs the best on PanoContext dataset and offers more robustness to occlusion from foreground objects. DuLa-Net v2 outperforms in two of three metrics for Stanford 2D-3D. To evaluate the performance on reconstructing general Manhattan layout shapes, we extend the Matterport3D dataset with general Manhattan layout annotations and introduce the {\datasetName} dataset. The annotations contain panoramas of both simple (\eg cuboid) and complex room shapes. We introduce two depth based evaluation metrics for evaluating the quality of reconstruction. 

Future work can be in three directions: (1) Relax Manhattan constraints to general layout. In real cases indoor layouts are more complex and could have non-Manhattan property like arch. One research direction is to study approaches that could generalize across Manhattan layouts and non-Manhattan ones with curve ceilings or walls. (2) Use additional depth and normal information. Our approach is based on a single RGB image only, and we can acquire rich geometric information from either predicted depth map from a single image, or captured depth maps from sensors. Incorporating depth features to both network predictions and the post-processing step could help for more accurate 3D layout reconstruction; (3) Extend to multi-view based 3D layout reconstruction. Reconstruction from a single image is difficult due to occlusions either from other layout boundaries or foreground objects. We can extend our approach for layout reconstruction from multiple images. Using multiple images can recover a more complete floor plan and scene layout, which has various applications such as virtual 3D room walk through for real estate.